\documentclass[10pt,twocolumn,letterpaper]{article}

\usepackage{iccv}
\usepackage{times}
\usepackage{epsfig}
\usepackage{graphicx}
\usepackage{amsmath}
\usepackage{amssymb}
\usepackage{array}
\usepackage[caption=false,font=normalsize,labelfont=sf,textfont=sf]{subfig}
\usepackage{textcomp}
\usepackage{stfloats}
\usepackage{url}
\usepackage{verbatim}
\usepackage{cite}
\usepackage{mathtools}
\usepackage{amsthm}
\usepackage{bbm}
\usepackage{dsfont}
\usepackage{multirow,multicol}
\usepackage[dvipsnames,table]{xcolor}     % colors
\usepackage{siunitx}
\usepackage{footmisc}
\usepackage{booktabs}
\usepackage[linesnumbered,ruled,vlined]{algorithm2e}
\usepackage{colortbl}
\usepackage{pifont}
\colorlet{LightGreen}{SpringGreen!60}
\SetKwInput{KwInput}{Input}                % Set the Input
\SetKwInput{KwOutput}{Output}              % set the Output

\SetCommentSty{mycommfont}

\usepackage[pagebackref=true,breaklinks=true,letterpaper=true,colorlinks,bookmarks=false]{hyperref}

\iccvfinalcopy % *** Uncomment this line for the final submission

\def\iccvPaperID{2} % *** Enter the ICCV Paper ID here

\ificcvfinal\fi

\begin{document}

%%%%%%%%% TITLE
\title{Boosting Semi-Supervised Learning\\ by bridging high and low-confidence predictions}

\author{Khanh-Binh~Nguyen\\
Department of Electrical and Computer Engineering\\
Sungkyunkwan University, South Korea\\
{\tt\small binhnk@skku.edu}
\and
Joon-Sung~Yang\\
School of Electrical and Electronic Engineering\\ and Department of Systems Semiconductor Engineering\\
Yonsei University, South Korea\\
{\tt\small js.yang@yonsei.ac.kr}
}

\maketitle
\ificcvfinal\thispagestyle{empty}\fi

%%%%%%%%% ABSTRACT
\begin{abstract}
Pseudo-labeling is a crucial technique in semi-supervised learning (SSL), where artificial labels are generated for unlabeled data by a trained model, allowing for the simultaneous training of labeled and unlabeled data in a supervised setting. 
However, several studies have identified three main issues with pseudo-labeling-based approaches. 
Firstly, these methods heavily rely on predictions from the trained model, which may not always be accurate, leading to a confirmation bias problem. 
Secondly, the trained model may be overfitted to easy-to-learn examples, ignoring hard-to-learn ones, resulting in the \textit{"Matthew effect"} where the already strong become stronger and the weak weaker.
Thirdly, most of the low-confidence predictions of unlabeled data are discarded due to the use of a high threshold, leading to an underutilization of unlabeled data during training. 
To address these issues, we propose a new method called ReFixMatch, which aims to utilize all of the unlabeled data during training, thus improving the generalizability of the model and performance on SSL benchmarks. 
Notably, ReFixMatch achieves 41.05\% top-1 accuracy with 100k labeled examples on ImageNet, outperforming the baseline FixMatch and current state-of-the-art methods.
\end{abstract}

%%%%%%%%% BODY TEXT
\section{Introduction}
The strengths of Deep Neural Networks (DNNs) have been proven through numerous successes in a wide range of tasks, such as image classification \cite{he2016deep}, speech recognition \cite{amodei2016deep}, and natural language processing \cite{socher2012deep}.
Despite the high performance and state-of-the-art benchmarks, the superior performance of DNNs heavily relies on training with a large amount of labeled data \cite{hestness2017deep,jozefowicz2016exploring,mahajan2018exploring,radford2019language,raffel2020exploring}.
In addition, there are also challenges in using large labeled datasets, such as the availability of the datasets, the cost of collecting and labeling data, etc.
To alleviate the dependence on labeled data, semi-supervised learning (SSL) has been proposed.
With the advantages of using a large volume of unlabeled data, SSL has become a powerful method for training models.
Furthermore, using SSL not only reduces the cost of collecting data but also produces equivalent results to supervised learning approaches.
This success has led to the development of many SSL methods \cite{berthelot2019mixmatch,Berthelot2020ReMixMatchSL,Laine2017TemporalEF,Lee2013PseudoLabelT,Tarvainen2017MeanTA,Xie2020SelfTrainingWN}. 
A popular approach of SSL methods is to produce an artificial label for unlabeled data and train the model using the artificial label as ground truth.
For example, the pseudo-labeling \cite{Lee2013PseudoLabelT} (categorized as self-training \cite{26629,Xie2020SelfTrainingWN} method) uses the model’s class prediction as a pseudo-label to train.
It is a well-established technique for semi-supervised learning \cite{liu2019deep,sohn2020fixmatch}, domain adaptation \cite{kang2019contrastive,na2021fixbi}, and transfer learning \cite{arnold2007comparative}. 
Unlike pseudo-labeling, consistency regularization uses loss functions such as mean squared error (MSE) or Kullback-Leibler divergence (KL divergence) to minimize the difference between model predictions for different augmented inputs.

Recent work from \cite{sohn2020fixmatch} suggests using a high threshold to filter out only reliable pseudo-labels for training and masking out the rest.
FlexMatch \cite{zhang2021flexmatch} improves the performance of FixMatch by applying the Curriculum Pseudo Labeling (CPL) method to let the model learn equally among classes with class-wise dynamic thresholds.
CoMatch \cite{li2021comatch} uses Contrastive Graph Regularization to improve performance by learning jointly-evolved class probabilities and image representations.
SimMatch \cite{zheng2022simmatch} simultaneously considers semantic similarity and instance similarity of the data.
While achieving state-of-the-art performance, FixMatch and its variants \cite{zhang2021flexmatch,li2021comatch,zheng2022simmatch,kuo2020featmatch} are still encountering the confirmation bias problem \cite{arazo2020pseudo}.
To eliminate the effects of learning on biased pseudo-label, a number of works have been proposed \cite{zheng2022simmatch,yao2022cls,xu2021dp,berthelot2021adamatch,hu2021simple}.
However, because of the high threshold setting, a large proportion of unlabeled data with prediction scores below the threshold is discarded during training and never used, especially for hard-to-learn classes.
This leads to another major issue that the unlabeled data is not fully exploited for FixMatch and many studies based on it.
Furthermore, to tackle the confirmation bias issue, the previous studies introduced additional modules and extra computational overhead.

\begin{figure}[!ht]
    \centering
    \subfloat[CIFAR-10\label{fig:mrcifar10}]
    {\includegraphics[width=.49\linewidth]{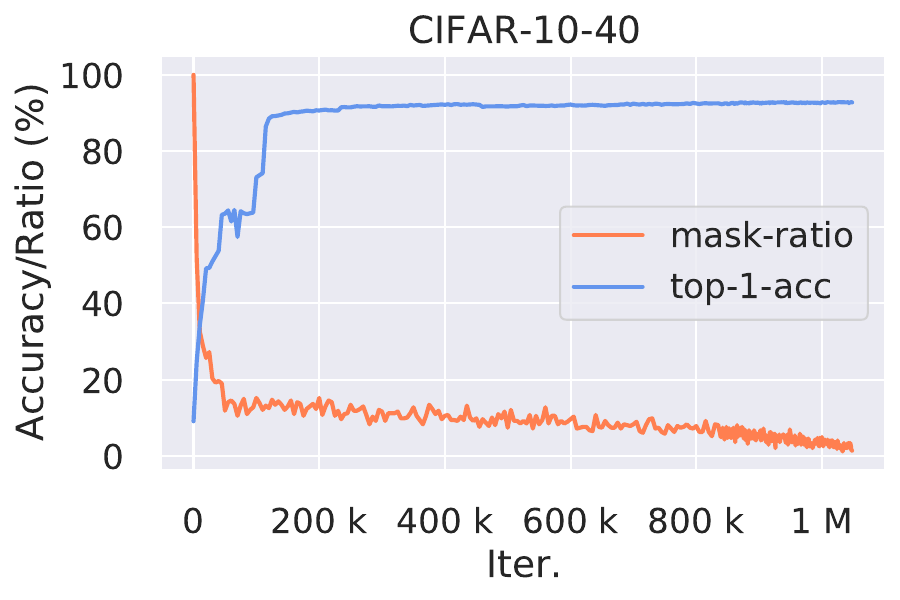}}%
    \subfloat[CIFAR-100\label{fig:mrcifar100}]
    {\includegraphics[width=.49\linewidth]{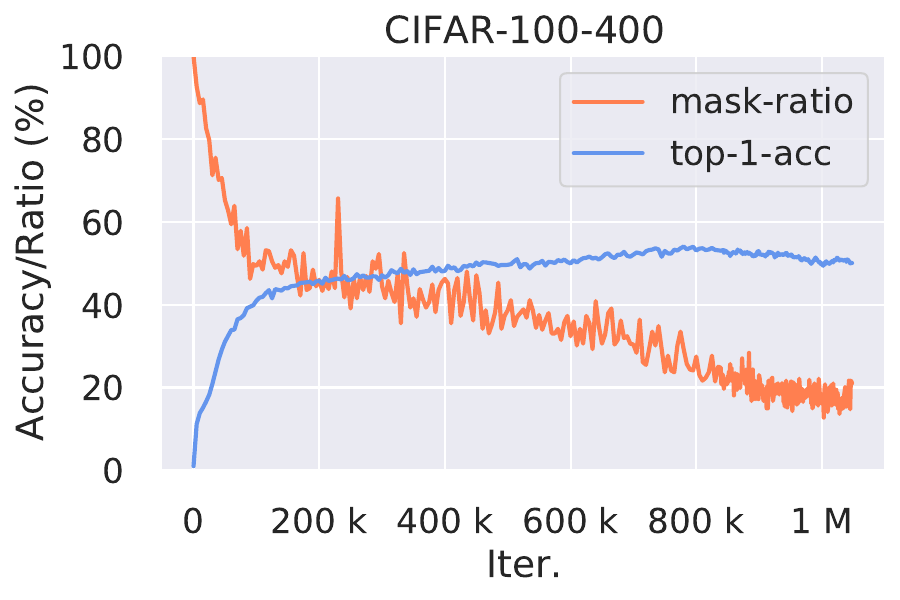}}
    \caption{Top-1 accuracy vs mask ratio of unlabeled data from FixMatch. \textbf{(a)} CIFAR-10 40-label split. \textbf{(b)} CIFAR-100 400-label split.}
    \label{fig:mask-ratio}
\end{figure}

We visualize the correlation between a top-1 accuracy and a mask ratio on CIFAR-10/100 in Figure \ref{fig:mask-ratio}.
It can be seen that while the number of qualified pseudo-labels is increasing by iterations, the accuracy just slowly increases and starts to decrease after 800k iterations.
This problem is clearly noticeable for large datasets such as CIFAR-100 in Figure \ref{fig:mrcifar100}.
Furthermore, the number of qualified pseudo-labels that are used during training only ranges from 60\% to 80\% of total unlabeled data in CIFAR-100.

In this work, we propose a simple SSL pipeline, ReFixMatch, which is shown in Figure \ref{fig:pipeline}.
Conventionally, UDA \cite{Xie2020UnsupervisedDA}, MixMatch \cite{berthelot2019mixmatch}, and ReMixMatch \cite{Berthelot2020ReMixMatchSL} train models with "soft" pseudo-labels for the whole unlabeled dataset.
Later, FixMatch \cite{sohn2020fixmatch} simplifies them by using only "hard" pseudo-labels from the high-confidence predictions.
FixMatch also shows that with the high-confidence threshold, sharpening the predictions into "soft" pseudo-labels does not lead to a significant difference in performance.
Hence, they discard the low-confidence predictions during training.
Unlike previous approaches, ReFixMatch aims to maximize the utilization of the whole unlabeled dataset to improve generalization during training.
Specifically, we bridge the usage of "hard" pseudo-labels from high-confidence predictions and "soft" pseudo-labels from low-confidence predictions.
Thus, the low-confidence predictions would be considered guesses, and the information from them could be transferred to the model to improve its performance and representation.
% , similar to transferring useful information in Knowledge Distillation.
In this manner, we leverage the advantages of both "hard" and "soft" pseudo-labels as well as the whole unlabeled dataset.
The use of low-confidence samples has already been well studied in many other related tasks. 
This usage, however, is still being studied for semi-supervised learning tasks. 
There are also research that leverages low-confidence predictions, such as \cite{feng2022dmt,zhao2022lassl}. 
However, in order to enhance the learning process, they either use multiple models or introduce a complicated pipeline. 
This work presents an efficient yet straightforward approach based on FixMatch, the most widely used SSL pipeline.
The novelty of ReFixMatch lies in the simplicity, which helps it outperform SOTA methods, which are much more complex. 
ReFixMatch adds no overhead to the conventional pipeline since it only uses an extra loss term. 
Because of this simplicity, many SSL frameworks, including semi-supervised semantic segmentation and object detection can be benefitted from this study.
The benefit of introducing ReFixMatch is particularly remarkable on all datasets, especially imbalanced datasets.
ReFixMatch achieves 28.60\%, 8.39\%, and 6.11\% error rates when the number of labels is 40, 250, and 1000, respectively, on the STL-10 dataset.
Furthermore, on the SVHN dataset, ReFixMatch achieves 2.15\% and 1.89\%hl error rate; ReFixMatch with CPL gives 2.63\% and 2.01\% error rates when the label amount is 40 and 1000, respectively, while FlexMatch fails with a large margin.
ReFixMatch also improves the convergence speed and the generalization of the model.

% \newpage
To sum up, this paper makes the following contributions:
\begin{itemize}
    \item We systematically investigate and analyze the importance of low-confidence predictions for unlabeled data in the training of SSL methods.
    \item We propose a simple yet effective method, ReFixMatch, to leverage the whole unlabeled data set, including high and low-confidence predictions.
    \item ReFixMatch introduces no additional modules or extra computational overhead, and it can be used with any SSL method to improve performance.
    \item ReFixMatch establishes a new state-of-the-art performance for semi-supervised learning. ReFixMatch achieves a 41.05\% error rate on ImageNet with 100k labeled images and outperforms prior methods.
\end{itemize}

\section{Analysis of high-confidence and low-confidence pseudo-label}
In order to examine the importance of low-confidence predictions in the training process, we train FixMatch separately with "hard" and "soft" pseudo-labels.
The "hard" pseudo-label training is the conventional FixMatch using high-confidence predictions, while for the "soft" pseudo-label training, the model is trained only on low-confidence predictions.
Specifically, instead of choosing high-confidence predictions as the pseudo-label, we take the low-confidence predictions from weakly-augmented examples, sharpen them by temperature $\mathbf{T=0.5}$ and compute the KL divergence with the predictions from strongly-augmented.

\begin{table}[!h]
    \caption{Error rate of FixMatch using high-confidence vs low-confidence predictions on CIFAR-10 with 40, 250, and 1000-label split.}
    \label{table:high-low}
    \centering
    \resizebox{\linewidth}{!}{%
    \begin{sc}
        \begin{tabular}{lcc} \toprule   \midrule
        Dataset             & high-confidence    & low-confidence     \\ \midrule   \midrule
        CIFAR-10-40         & 7.47          & 28.88         \\
        CIFAR-10-250        & 4.86          & 8.07          \\
        CIFAR-10-4000       & 4.21          & 8.04          \\ \midrule \bottomrule
        \end{tabular}
    \end{sc}
    }
\end{table}

The experiment results from Table \ref{table:high-low} show that using only low-confidence predictions to train the model can still achieve a competitive performance with the one using high-confidence predictions on the CIFAR-10 dataset.
This shows that the conventional approach of using a high threshold and discarding a large proportion of unlabeled data during training is inefficient and does not fully leverage the unlabeled data.
Thus, instead of using only high-confidence predictions, in this work, we bridge the strengths of both high-confidence and low-confidence predictions.

\section{ReFixMatch}
We propose ReFixMatch, a simple SSL pipeline that considers the information from the whole unlabeled dataset.
The main novelty of ReFixMatch is the utilization of unlabeled examples that have a prediction probability lower than the threshold $\tau$.
In the following section, we explain the whole process of ReFixMatch for semi-supervised image classification problems.

\subsection{ReFixMatch pipeline}
\begin{figure}[ht!]
    \centering
    \resizebox{\linewidth}{!}{\includegraphics{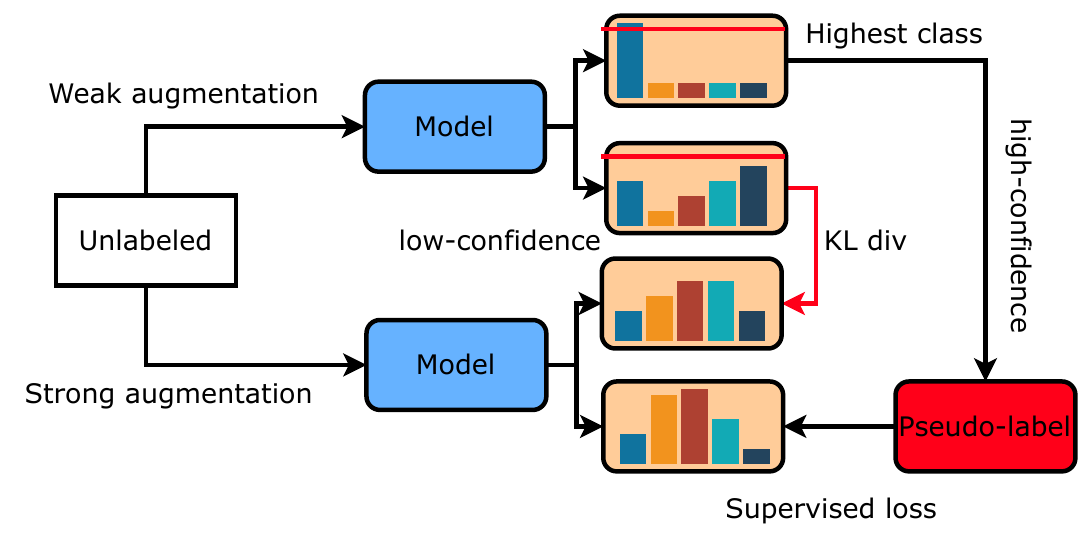}}
    \caption{Diagram of the proposed ReFixMatch.
    For weakly-augmented predictions, the high-confidence predictions are converted to one-hot pseudo-label, while the low-confidence predictions are sharpened with temperature $\mathbf{T}$.
    We measure the Kullback-Leibler divergence loss for the sharpened low-confidence predictions with the strongly-augmented predictions from the same input and the cross-entropy loss for the "hard" pseudo-label.
    }
    \label{fig:pipeline}
\end{figure}

Our proposed ReFixMatch pipeline consists of two phases, as shown in Figure \ref{fig:pipeline}.
In the training phase, we perform supervised training for the model with labeled data and evaluate the standard cross-entropy loss.
During the inference phase, two perturbed versions of unlabeled images, which are either weakly or strongly augmented, are created.
Then, for the unlabeled data, pseudo-labels are generated from the high-confidence predictions of the weakly-augmented unlabeled version.
Next, these pseudo-labels are used to supervise the model prediction of the strongly-augmented version on the next iteration, together with labeled data.
Last, we sharpen the low-confidence predictions of the weakly-augmented unlabeled version.
A KL divergence loss function is used for the sharpened low-confidence predictions and the predictions from the strongly-augmented version.

While minimizing the cross entropy loss between model logits and hard one-hot targets remains the go-to recipe for supervised  classification  training,  learning  from  soft  tar-get  emerges  in  many  lines  of  research.    
Label Smoothing  [36, 43]  is  a  straightforward  method  that  applies  a fixed  smoothing  (softening)  factor $\alpha$ to  the  hard  one-hot classification target. 
The motivation is that label smoothing prevents the model from becoming over-confident.

\subsection{Preliminaries}
In SSL, the training data consists of labeled and unlabeled data. 
Let $\mathcal{X}=\left\{\left(x_{b}, y_{b}\right): b \in(1, \ldots, B)\right\}$ be a batch of $B$ labeled examples, where $x_b$ is training examples and $y_b$ is one-hot labels, and $\mathcal{U}=\left\{u_{b}: b \in \left(1,\dots,\mu{B}\right)\right\}$ be a batch of $\mu{B}$ unlabeled examples where $\mu$ is a hyperparameter determining the relative sizes of $\mathcal{X}$ and $\mathcal{U}$. 

We construct the loss function of our proposed ReFixMatch with the supervised loss, which is a standard cross-entropy loss ($\mathcal{L}_{s}^\mathrm{CE}$) for the labeled data, and the unsupervised loss, including the KL divergence loss ($\mathcal{L}_\mathrm{KL}$) for the low-confidence predictions and the standard cross-entropy loss ($\mathcal{L}_{u}^\mathrm{CE}$) for high-confidence predictions.
\begin{equation}
    \mathcal{L}_\mathrm{SSL}=\mathcal{L}_{s}^\mathrm{CE} + \lambda_u\mathcal{L}_{u}
\end{equation}
where $\lambda_u$ is the fixed weight for the unlabeled data loss.

Specifically, $\mathcal{L}_{s}^\mathrm{CE}$ is a standard cross-entropy loss on weakly-augmented labeled data:
\begin{equation}
    \mathcal{L}_{s}^\mathrm{CE}=\frac{1}{B} \sum_{b=1}^{B} \mathrm{H}\left(y_{b}, p_{m}\left(y \mid \mathcal{A}_w\left(x_{b}\right)\right)\right)
\end{equation}
where $p_m(y|x)$ is the predicted class distribution of the model for input $x$, and $H(p,q)$ denotes the "hard" label cross-entropy between two probability distributions, $p$ and $q$.
Then, let $\mathcal{A}_w$ be the weakly (i.e., random crop and flip) augmentation and $\mathcal{A}_s$ be the strongly (i.e., RandAugment \cite{Cubuk2020RandaugmentPA}) augmentation for unlabeled data, respectively.
$\mathcal{L}_{u}$ is defined as a total of the standard cross-entropy loss ($\mathcal{L}_{u}^\mathrm{CE}$) and the KL divergence loss ($\mathcal{L}_\mathrm{KL}$).
$\mathcal{L}_{u}^\mathrm{CE}$ is the loss between the high-confidence pseudo-label of weakly-augmented unlabeled data and the predictions of the model for strongly-augmented unlabeled data.
$\mathcal{L}_\mathrm{KL}$ is the KL divergence loss between the sharpened low-confidence predictions of weakly-augmented examples $\mathcal{A}_w$ and the predictions of strongly-augmented examples $\mathcal{A}_s$, defined as:
\begin{equation}
    \mathcal{L}_{u}=\mathcal{L}_{u}^\mathrm{CE}+\mathcal{L}_\mathrm{KL}
\end{equation}
\begin{equation}
    \mathcal{L}_{u}^\mathrm{CE}=\frac{1}{\mu B}\sum_{b=1}^{\mu B} \mathbbm{1}\left(\max \left(q_{b}\right) \geq \tau\right) \mathrm{H}\left(\hat{q}_{b}, p_{m}\left(y \mid \mathcal{A}_{s}\left(u_{b}\right)\right)\right)
\end{equation}
\begin{equation}
    \footnotesize{\mathcal{L}_\mathrm{KL} = \frac{1}{\mu B}\sum_{b=1}^{\mu B}\mathbbm{1}\left(\max \left(q_{b}\right)<\tau\right)D_\mathrm{KL}\left(p_{s}^{w} \mid  p_{m}\left(y \mid \mathcal{A}_{s}\left(u_{b}\right)\right)\right)}
\end{equation}
\begin{equation}
    p_{s}^{w}\left(y \mid \mathcal{A}_w\left(u_{b}\right)\right) = \frac{\exp{\left(z_{b}/\mathbf{T}\right)}}{\sum_{k}\exp{\left(z_{k}/\mathbf{T}\right)}}
\end{equation}
where $\hat{q}_{b}=\arg\max\left(q_b\right)$ is the pseudo-label with $q_b=p_m\left(y \mid \mathcal{A}_w\left(u_b\right)\right)$ for input $\mathcal{A}_w\left(u_b\right)$, $\tau$ is the threshold for choosing pseudo-label, $D_\mathrm{KL}$ denotes the KL divergence function, $z_{b}$ is the logits for example, $\mathcal{A}_w\left(u_{b}\right)$ and $\mathbf{T}$ is the temperature for sharpening.

\subsection{Algorithm}
The algorithm for ReFixMatch is presented in Algorithm \ref{alg:cap}.
Compared to FlexMatch, the ReFixMatch algorithm is much simpler, as it does not require computation of the threshold for each iteration.
The algorithm of ReFixMatch is as simple as FixMatch, with only additional loss for low-confidence predictions.
Therefore, ReFixMatch does not require any additional budget compared to prior methods.

\begin{algorithm}[!ht]
\DontPrintSemicolon
  
    \KwInput{Labeled batch $\mathcal{X}={\left(x_b,y_b\right):b\in{\left(1,\dots,B\right)}}$, unlabeled batch $\mathcal{U}={u_b:b\in{\left(1,\dots,\mu B\right)}}$, confidence threshold $\tau$, unlabeled data ratio $\mu$, unlabeled loss weight $\lambda_u$, temperature $\mathbf{T}$}
    $\mathcal{L}_{s}^\mathrm{CE}=\frac{1}{B} \sum_{b=1}^{B} \mathrm{H}\left(y_{b}, p_{m}\left(y \mid \mathcal{A}_w\left(x_{b}\right)\right)\right)$
    \tcp*[l]{Cross-entropy loss for labeled data}
    \For{$b = 1$ \KwTo $\mu B$}{
        \tcc{Compute prediction after applying weak data augmentation of $u_b$}
        $q_b = p_m\left(y\mid\mathcal{A}_w\left(u_b\right);\theta\right)$
        
        \tcc{Sharpen the low-confidence predictions}
        $p_{s}^{w}\left(y \mid \mathcal{A}_w\left(u_{b}\right)\right) = \frac{\exp{\left(z_{b}/\mathbf{T}\right)}}{\sum_{k}\exp{\left(z_{k}/\mathbf{T}\right)}}$
        
    }
    \tcc{Cross-entropy loss with pseudo-label and confidence threshold for high-confidence unlabeled data}
    $\mathcal{L}_{u}^{CE}=\frac{1}{\mu B}\left(\sum_{b=1}^{\mu B} \mathbbm{1}\left(\max \left(q_{b}\right) \geq \tau\right)\mathrm{H}\left(\hat{q}_{b}, p_{m}\left(y \mid \mathcal{A}_{s}\left(u_{b}\right)\right)\right)\right)$
    
    \tcc{Kullback-Leibler divergence loss with sharpened pseudo-label and confidence threshold for low-confidence unlabeled data}
    $\mathcal{L}_{KL} = \frac{1}{\mu B}\left(\sum_{b=1}^{\mu B}\mathbbm{1}\left(\max \left(q_{b}\right)<\tau\right)\right.$ $\left.D_\mathrm{KL}\left(p_{s}^{w} \mid  p_{m}\left(y \mid \mathcal{A}_{s}\left(u_{b}\right)\right)\right)\right)$\\
    
    $\mathcal{L}_u = \mathcal{L}_u^{CE} + \mathcal{L}_{KL}$
    
    \Return{$\mathcal{L}_s^{CE}+\lambda_{u} \mathcal{L}_u$}
\caption{ReFixMatch algorithm}
\label{alg:cap}
\end{algorithm}

\section{Experiments}
We evaluate ReFixMatch on common datasets such as CIFAR-10/100 \cite{krizhevsky2009learning}, SVHN \cite{netzer2011reading}, STL-10 \cite{coates2011analysis}, and ImageNet \cite{Deng2009ImageNetAL} under various labeled data amounts. 
We mainly compare our proposed method with recent state-of-the-art methods such as UDA \cite{Xie2020UnsupervisedDA}, FixMatch \cite{sohn2020fixmatch}, FlexMatch \cite{zhang2021flexmatch}, CoMatch \cite{li2021comatch}, SimMatch \cite{zheng2022simmatch}, and AdaMatch \cite{berthelot2021adamatch}.
We also include a fully-supervised experiment for each dataset.
The implementation and evaluation of all methods are based on TorchSSL\footnote{https://github.com/TorchSSL/TorchSSL}.

We use the same training hyperparameters for a fair comparison of UDA, FixMatch, and FlexMatch methods.
There are only minor differences for some hyperparameters regarding each method algorithm settings.
Standard stochastic gradient descent (SGD) with a momentum of 0.9 is used as an optimizer in all experiments \cite{sutskever2013importance,polyak1964some}. 
An initial learning rate of 0.03 with a cosine annealing learning rate scheduler \cite{Loshchilov2017SGDRSG} is used for a total of $2^{20}$ training iterations.
We also conducted an exponential moving average with a momentum of 0.999. 
The batch size of labeled data is 64 for all datasets except ImageNet. 
$\mu$ is set to 7 for CIFAR-10/100, SVHN and STL-10, and it is set to 1 for ImageNet.
$\tau$ is set to 0.8 for UDA and is set to 0.95 for FixMatch, FlexMatch, and ReFixMatch.
These configurations follow the original papers \cite{Xie2020UnsupervisedDA,sohn2020fixmatch,zhang2021flexmatch}.
We set $\mathbf{T}$ to 0.4 for UDA and 0.5 for ReFixMatch.
The strong augmentation in our experiments is RandAugment \cite{Cubuk2020RandaugmentPA}. 
For the ImageNet dataset, we use ResNet-50 \cite{krizhevsky2009learning} and for other datasets, we use variants of Wide-ResNet (WRN).

\subsection{CIFAR-10/100, STL-10, SVHN}
\label{sec:cifar}
We evaluate the best error rate by averaging the results of five runs with different random seeds for each method.
The classification error rates on the CIFAR-10/100, STL-10, and SVHN datasets are given in Table \ref{table:results1}.

\begin{table*}[ht!]
\centering
\caption{Error rates on CIFAR-10/100, SVHN, and STL-10 datasets on 5 different folds. 
All models are tested using the same code base from TorchSSL. \textbf{Bold} indicates best result and \underline{Underline} indicates second-best result.}
\label{table:results1}
\resizebox{\textwidth}{!}{%
\begin{sc}
\begin{tabular}{l|ccc|ccc|ccc|cc}
\toprule   \midrule
Dataset & \multicolumn{3}{c|}{CIFAR-10} & \multicolumn{3}{c|}{CIFAR-100} & \multicolumn{3}{c|}{STL-10} & \multicolumn{2}{c}{SVHN} \\ \midrule
Label Amount & 40 & 250 & 4000 & 400 & 2500 & 10000 & 40 & 250 & 1000 & 40 & 1000 \\ \midrule   \midrule
UDA \cite{Xie2020UnsupervisedDA} & 10.62\tiny{$\pm$3.75} & 5.16\tiny{$\pm$0.06} & 4.29\tiny{$\pm$0.07} & 46.39\tiny{$\pm$1.59} & 27.73\tiny{$\pm$0.21} & 22.49\tiny{$\pm$0.23} & 37.42\tiny{$\pm$8.44} & 9.72\tiny{$\pm$1.15} & 6.64\tiny{$\pm$0.17} & 5.12\tiny{$\pm$4.27} & \underline{1.89\tiny{$\pm$0.01}} \\
FixMatch \cite{sohn2020fixmatch} & 7.47\tiny{$\pm$0.28} & 4.86\tiny{$\pm$0.05} & 4.21\tiny{$\pm$0.08} & 46.42\tiny{$\pm$0.82} & 28.03\tiny{$\pm$0.16} & 22.20\tiny{$\pm$0.12} & 35.97\tiny{$\pm$4.14} & 9.81\tiny{$\pm$1.04} & 6.25\tiny{$\pm$0.33} & 3.81\tiny{$\pm$1.18} & 1.96\tiny{$\pm$0.03} \\
FlexMatch \cite{zhang2021flexmatch} & 4.97\tiny{$\pm$0.06} & 4.98\tiny{$\pm$0.09} & 4.19\tiny{$\pm$0.01} & 39.94\tiny{$\pm$1.62} & \underline{26.49\tiny{$\pm$0.20}} & 21.90\tiny{$\pm$0.15} & 29.15\tiny{$\pm$4.16} & 8.23\tiny{$\pm$0.39} & 5.77\tiny{$\pm$0.18} & 8.19\tiny{$\pm$3.20} & 6.72\tiny{$\pm$0.30} \\
CoMatch \cite{li2021comatch} & 6.51\tiny{$\pm$1.18} & 5.35\tiny{$\pm$0.14} & 4.27\tiny{$\pm$0.12} & 53.41\tiny{$\pm$2.36} & 29.78\tiny{$\pm$0.11} & 22.11\tiny{$\pm$0.22} & \textbf{13.74\tiny{$\pm$4.20}} & \textbf{7.63\tiny{$\pm$0.94}} & \textbf{5.71\tiny{$\pm$0.08}} & 8.20\tiny{$\pm$5.32} & 2.01\tiny{$\pm$0.04} \\
SimMatch \cite{zheng2022simmatch} & 5.38\tiny{$\pm$0.01} & 5.36\tiny{$\pm$0.08} & 4.41\tiny{$\pm$0.07} & \underline{39.32\tiny{$\pm$0.72}} & \textbf{26.21\tiny{$\pm$0.37}} & \textbf{21.50\tiny{$\pm$0.11}} & \underline{16.98\tiny{$\pm$4.24}} & 8.27\tiny{$\pm$0.04} & 5.74\tiny{$\pm$0.31} & 7.60\tiny{$\pm$2.11} & 2.05\tiny{$\pm$0.05} \\
AdaMatch \cite{berthelot2021adamatch} & 5.09\tiny{$\pm$0.21} & 5.13\tiny{$\pm$0.05} & 4.36\tiny{$\pm$0.05} & \textbf{38.08\tiny{$\pm$1.35}} & 26.66\tiny{$\pm$0.33} & 21.99\tiny{$\pm$0.15} & 19.95\tiny{$\pm$5.17} & 8.59\tiny{$\pm$0.43} & 6.01\tiny{$\pm$0.02} & 6.14\tiny{$\pm$5.35} & 2.02\tiny{$\pm$0.05} \\
\midrule
\rowcolor{LightGreen}\textbf{ReFixMatch} & \textbf{4.94\tiny{$\pm$0.01}} & \textbf{4.83\tiny{$\pm$0.05}} & \underline{4.18\tiny{$\pm$0.05}} & 46.12\tiny{$\pm$1.07} & 27.28\tiny{$\pm$0.22} & \underline{21.60\tiny{$\pm$0.04}} & 28.60\tiny{$\pm$4.21} & \underline{8.21\tiny{$\pm$0.30}} & \underline{5.74\tiny{$\pm$0.30}} & \textbf{2.15\tiny{$\pm$1.23}} & \textbf{1.89\tiny{$\pm$0.03}} \\
\rowcolor{LightGreen}\textbf{ReFixMatch + CPL} & \underline{4.95\tiny{$\pm$0.05}} & \underline{4.85\tiny{$\pm$0.06}} & \textbf{4.13\tiny{$\pm$0.02}} & 46.73\tiny{$\pm$1.37} & 27.25\tiny{$\pm$0.25} & 21.78\tiny{$\pm$0.04} & 28.66\tiny{$\pm$4.40} & 8.23\tiny{$\pm$0.31} & 5.76\tiny{$\pm$0.42} & \underline{2.63\tiny{$\pm$1.46}} & 2.01\tiny{$\pm$0.05} \\
\midrule
Fully-Supervised & \multicolumn{3}{c|}{4.62\tiny{$\pm$0.05}} & \multicolumn{3}{c|}{19.30\tiny{$\pm$0.09}} & \multicolumn{3}{c|}{None} & \multicolumn{2}{c}{2.13\tiny{$\pm$0.02}} \\ \midrule
\bottomrule
\end{tabular}%
\end{sc}
}
\end{table*}

We employ Wide-ResNet \cite{Zagoruyko2016WideRN} as a backbone model for experiments.
Detailed model selection is reported in Appendix \ref{sec:app}.
ReFixMatch achieves the best performance on most of the datasets with different amounts of labels, as shown in Table \ref{table:results1}.
ReFixMatch not only achieves high performance across all datasets but also performs well on the SVHN dataset, while FlexMatch performs less favorably on imbalanced datasets such as the SVHN \cite{zhang2021flexmatch}.
Especially, ReFixMatch using CPL improves the results of FlexMatch on the SVHN dataset.
This proves that our proposed method with the strategy of leveraging the whole unlabeled dataset can mitigate the overfitting issue when training on small and imbalanced datasets.
However, since ReFixMatch and FlexMatch have the same approach in common that helps the model utilize more data, using CPL with our proposed ReFixMatch results in a degradation of performance on balanced datasets.
Moreover, CPL improves the number of "hard" pseudo-labels, thus reducing the effect of ReFixMatch.
It should be noted that ReFixMatch adds no overhead, while CoMatch and SimMatch use additional complex modules.
They also use the distribution alignment technique, which provides much better results.

\subsection{ImageNet}
We further evaluate ReFixMatch on large and complex datasets such as ImageNet \cite{Deng2009ImageNetAL}.
We train the models with 100k of labeled data.
Furthermore, because the ImageNet dataset is large and complex, we set the $\tau$ threshold value to 0.7 to improve the capture of samples with the correct pseudo-label.
The batch size is 128 and the weight decay is 0.0003 for the 100k labels experiment.
For 10\% experiments, we follow the settings in \cite{sohn2020fixmatch,li2021comatch,zheng2022simmatch}.

\begin{table}[!ht]
\centering
    \caption{Error rate results on ImageNet.}
    \resizebox{0.8\linewidth}{!}{%
\begin{sc}
    \begin{tabular}{@{}l|c|c|c|c@{}}
    \toprule    \midrule
    \multirow{2}{*}{Method}                  & Top-1 & Top-5 & Top-1 & Top-5\\ 
    \cmidrule{2-5}
    & \multicolumn{2}{c|}{100k} & \multicolumn{2}{c}{10\%}\\
    \midrule    \midrule
    FixMatch \cite{sohn2020fixmatch}                    & 43.66 & 21.80 & 28.50 & 10.90  \\
    FlexMatch \cite{zhang2021flexmatch}                 & 41.85 & 19.48 & - & - \\
    CoMatch \cite{li2021comatch}                        & 42.17 & 19.64 & 26.30 & 8.60 \\
    SimMatch \cite{zheng2022simmatch}                   & - & - & 25.60 & 8.40 \\
    \midrule
    \rowcolor{LightGreen}\textbf{ReFixMatch}           & \textbf{41.05} & \textbf{19.01} & \textbf{24.80} & \textbf{8.10} \\ 
    \rowcolor{LightGreen}\textbf{ReFixMatch+CPL}           & 41.75 & 19.36 & - & - \\ \midrule    \bottomrule
    \end{tabular}%
\end{sc}
    }
    \label{table:results2}
\end{table}

As reported in Table \ref{table:results2}, ReFixMatch outperforms FixMatch, FlexMatch, and CoMatch with 41.05\% and 19.01\% for the top-1 and top-5 error rates, respectively.
This result clearly indicates that our proposed ReFixMatch can help boost performance for large and complex datasets such as ImageNet, especially when they are imbalanced (the number of images per class in the ImageNet dataset ranges between 732 and 1300).
Besides, when applying CPL from FlexMatch \cite{zhang2021flexmatch} to ReFixMatch, the results drop to 41.75\% and 19.36\% for the top-1 and top-5 error rates, respectively, as we explained in Section \ref{sec:cifar}.
In addition, ReFixMatch also surpasses the best performance of CoMatch and SimMatch by a large margin with 10\% labels; details are in Appendix \ref{app:details}.
% To further verify the strengths of ReFixMatch, we conduct comparisons on the same settings of SimMatch \cite{zheng2022simmatch} and CoMatch \cite{li2021comatch}.
% We use the source code from SimMatch\footnote{https://github.com/KyleZheng1997/simmatch} and CoMatch\footnote{https://github.com/salesforce/CoMatch}, and report the results in Table \ref{table:results3b}.

\subsection{Ablation Study}
\subsubsection{Training Efficiency}
\begin{figure}[!ht]
    \centering
    \subfloat[\footnotesize{Top-1 accuracy}]
    {\includegraphics[width=.49\linewidth]{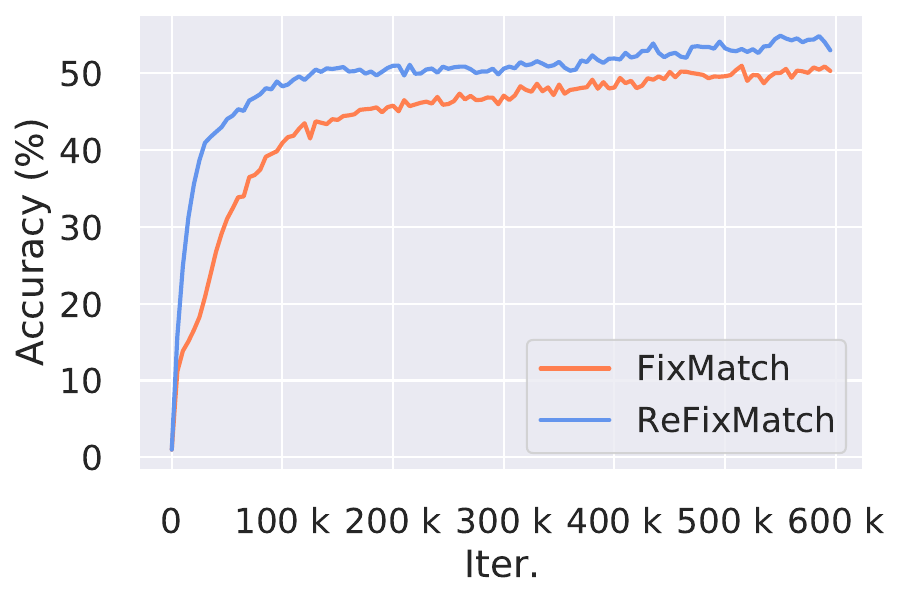}}%
    \subfloat[\footnotesize{Eval loss}]
    {\includegraphics[width=.49\linewidth]{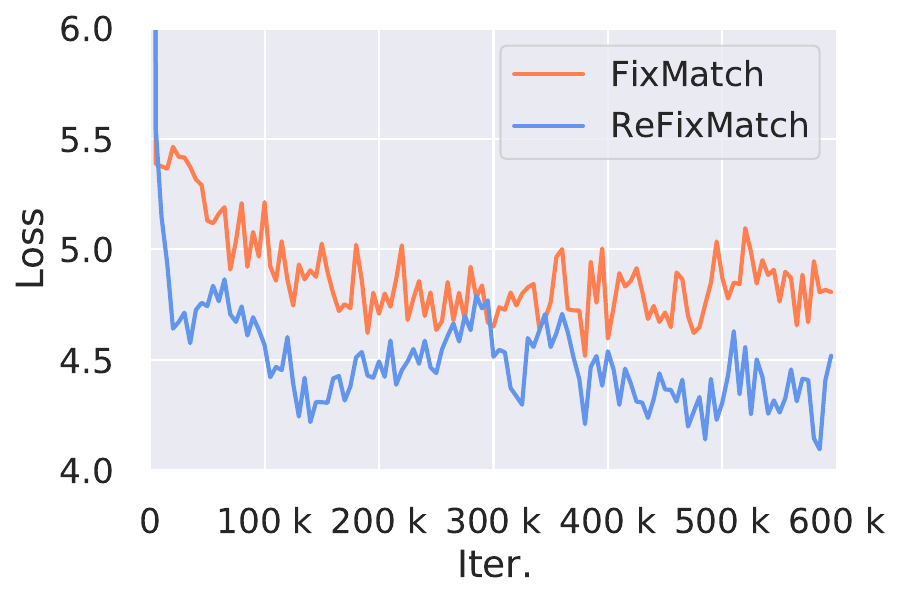}}
    \caption{Convergence analysis of ReFixMatch and FixMatch. \textbf{(a)}, \textbf{(b)} depict top-1 accuracy and loss on CIFAR-100 with 400 labels.}
    \label{fig:loss_acc}
\end{figure}

The convergence speed of our proposed ReFixMatch is extremely noticeable through our extensive experiments.
As we can see in Figure \ref{fig:loss_acc}, on CIFAR-100, ReFixMatch achieves over 40\% of accuracy within the first few iterations, while FixMatch nearly hits 20\%.
After 200k iterations, ReFixMatch achieves over 50\% accuracy, while FixMatch only achieves around 45\% of accuracy.
Moreover, the loss landscape of our proposed ReFixMatch also decreases faster than that of FixMatch.
In Figure \ref{fig:loss_acc}, we visualize the validation loss and top-1 accuracy of both FixMatch and ReFixMatch on the CIFAR-100 dataset with a 400-label split over 600k iterations for a better view of the difference.

\subsubsection{Class-wise accuracy on CIFAR-10 40-label split}
In Table \ref{table:classwise}, we present a thorough comparison of class-wise accuracy.
Our proposed ReFixMatch maintains high accuracy in easy-to-learn classes while simultaneously improving the accuracy in hard-to-learn classes.
ReFixMatch's final class-wise accuracy is balanced across classes, including hard-to-learn classes.
This demonstrates that employing both high and low-confidence predictions enhances not just the overall performance of the trained model but also the performance of each class.
ReFixMatch clearly outperforms FixMatch in class-wise accuracy in the evaluation phase for hard-to-learn classes.

\begin{table*}[!ht]
\centering
\caption{Class-wise accuracy comparison on CIFAR-10 40-label split.}
\label{table:classwise}
\resizebox{0.8\textwidth}{!}{%
\begin{sc}
\begin{tabular}{@{}l|cccccccccc@{}}
\toprule \midrule
Class Number    & 0 & 1 & 2 & 3 & 4 & 5 & 6 & 7 & 8 & 9\\ 
\midrule    \midrule
FixMatch        & 0.964 & 0.982 & 0.697 & 0.852 & 0.974 & \textbf{0.890} & \textbf{0.987} & 0.970 & 0.982 & \textbf{0.981}\\
\rowcolor{LightGreen}\textbf{ReFixMatch}   & \textbf{0.971} & \textbf{0.984} & \textbf{0.905} & \textbf{0.881} & \textbf{0.977} & 0.872 & 0.984 & \textbf{0.974} & \textbf{0.984} & 0.98\\ \midrule
FlexMatch       & 0.967 & 0.980 & \textbf{0.921} & 0.866 & 0.957 & 0.883 & \textbf{0.988} & \textbf{0.975} & 0.982 & 0.968\\
\rowcolor{LightGreen}\textbf{ReFixMatch + CPL\cite{zhang2021flexmatch}} & \textbf{0.967} & \textbf{0.983} & 0.915 & \textbf{0.876} & \textbf{0.969} & \textbf{0.889} & 0.971 & 0.974 & \textbf{0.985} & \textbf{0.973}\\ 
\midrule    \bottomrule
\end{tabular}%
\end{sc}
}
\end{table*}

\begin{figure}[!ht]
    \centering
    \includegraphics[width=\linewidth]{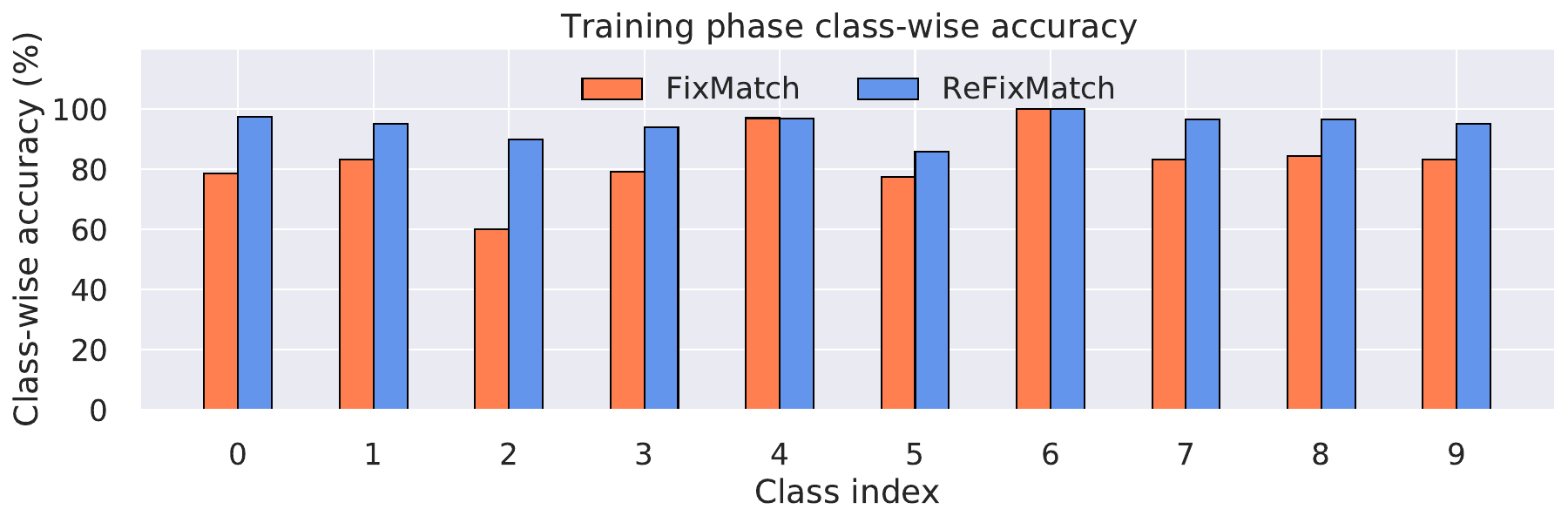}
    \includegraphics[width=\linewidth]{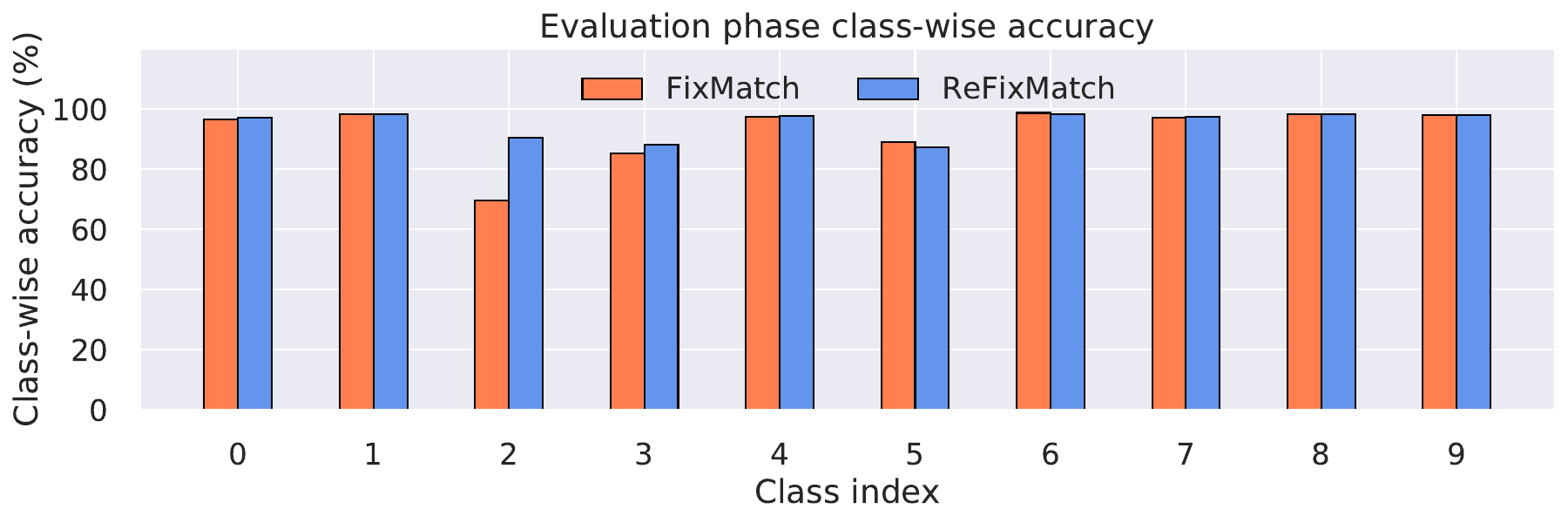}
    \caption{Class-wise accuracy comparison on CIFAR-10 40-label split at the best iteration of FixMatch and ReFixMatch.}
    \label{fig:classwise}
\end{figure}

The class-wise accuracy from the training phase, as shown in Figure \ref{fig:classwise}, indicates that leveraging the whole unlabeled dataset can improve the generalization of the model.
The results show that our ReFixMatch class-wise accuracy is much higher than FixMatch, and it also is balanced between easy-to-learn and hard-to-learn classes.

\begin{figure}[ht]
    \centering
    \includegraphics[width=0.5\linewidth]{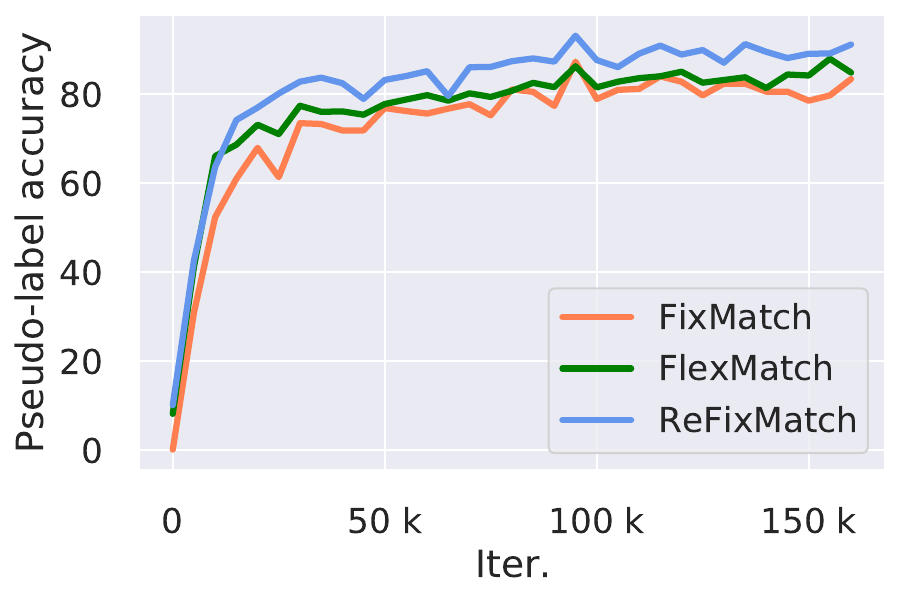}
    \caption{Pseudo-label accuracy.}
    \label{fig:pseudo-acc}
\end{figure}

Figure \ref{fig:pseudo-acc} shows the accuracy of the pseudo-label during training on the CIFAR-10 40-label split.
We can see that ReFixMatch can improve the accuracy of the pseudo-label over both FixMatch and FlexMatch.

\subsubsection{Data utilization and mask ratio}
\begin{figure}[!ht]
    \centering
    \subfloat[\footnotesize{Mask ratio}\label{fig:mask}]
    {\includegraphics[width=.49\linewidth]{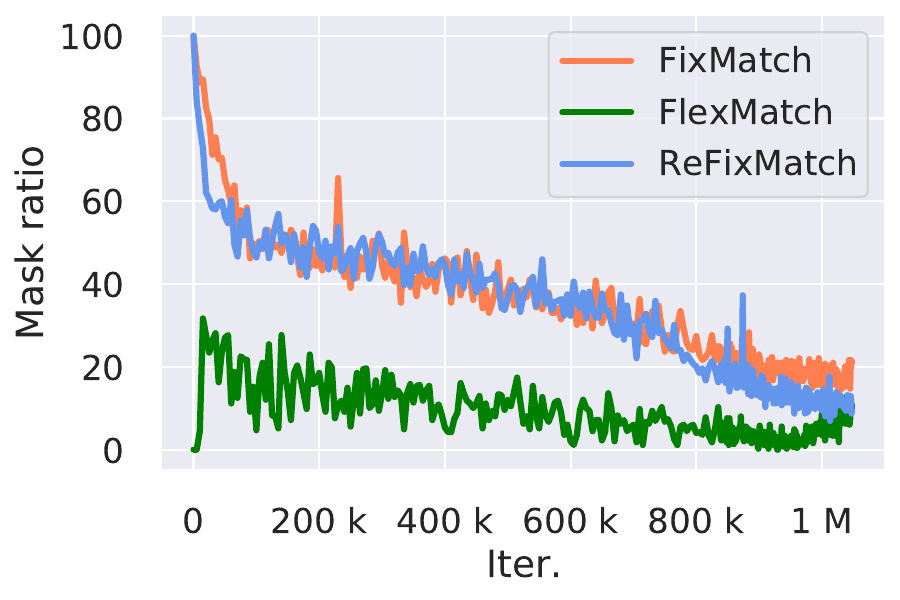}}%
    \subfloat[\footnotesize{Unlabeled data utilization}\label{fig:utilization}]
    {\includegraphics[width=.49\linewidth]{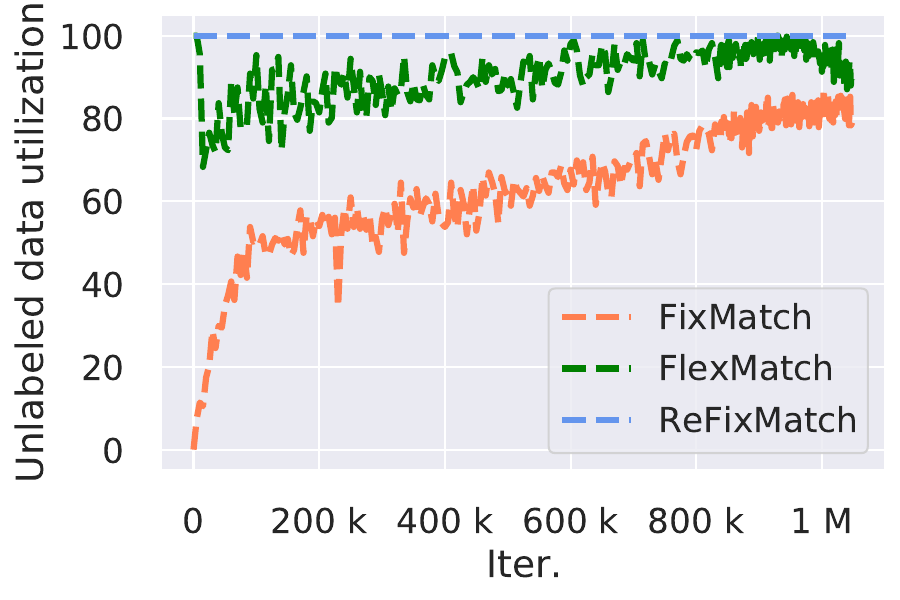}}
    \caption{Unlabeled data utilization and mask ratio on CIFAR-100 dataset with 400-label split.}
    \label{fig:mask-utilization}
\end{figure}

We present the unlabeled data utilization and mask ratio of FixMatch and ReFixMatch on the CIFAR-100 dataset with a 400-label split in Figures \ref{fig:mask}, \ref{fig:utilization}.
ReFixMatch helps reduce the mask-out data ratio and always uses the whole unlabeled dataset during training.
It also can be seen that the mask ratio of ReFixMatch less fluctuates than FixMatch.
It should be noted that FlexMatch has a lower mask ratio since it uses a lower threshold for each class, which allows the more low-confidence prediction to be used as pseudo-label but also introduces more noise to the model.

\subsubsection{CIFAR-10 Confusion Matrix}
\begin{figure}[!ht]
    \centering
    \includegraphics[width=\linewidth]{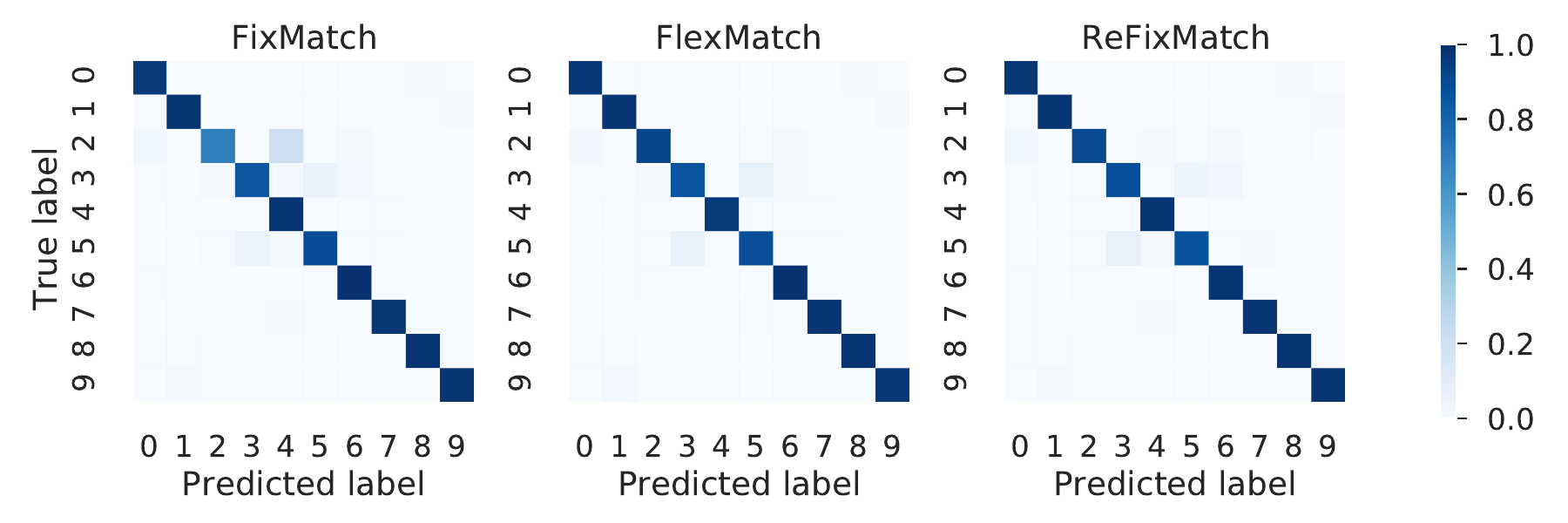}
    \caption{Confusion matrix of FixMatch, FlexMatch, and ReFixMatch features on CIFAR-10 with the 40-label split.}
    \label{fig:matrix}
\end{figure}

Figure \ref{fig:matrix} shows the confusion matrix of FixMatch, FlexMatch, and ReFixMatch on the CIFAR-10 dataset with a 40-label split.

\subsection*{Precision, Recall, F1 and AUC}
We also report precision, recall, F1-score, and AUC (area under curve) results on SVHN and STL-10 datasets with 40 labels to completely evaluate the performance of all methods in a classification setting.
As demonstrated in Table \ref{table:details}, ReFixMatch has the best performance in accuracy, recall, F1-score, and AUC, while also having lower error rates.
These measurements, along with error rates (accuracy), demonstrate the robust performance of our proposed method.
Especially on STL-10, simple ReFixMatch improves precision and recall by a large margin compared with prior methods.

\begin{table*}[ht!]
\centering
    \caption{Precision, recall, F1-score and AUC results on SVHN and STL-10.}
    \label{table:details}
\resizebox{0.8\textwidth}{!}{
\begin{sc}
    \begin{tabular}{@{}l|cccc|cccc}
    \toprule    \midrule
    Dataset    & \multicolumn{4}{c}{SVHN-40}      & \multicolumn{4}{|c}{STL-10-40}  \\ \midrule
    Criteria    & Precision & Recall    & F1 Score  & AUC       & Precision & Recall    & F1 Score  & AUC \\ \midrule   \midrule
    UDA \cite{Xie2020UnsupervisedDA}         & \textbf{0.9783}    & 0.9776    & 0.9777    & 0.9977    & 0.6385    & 0.5319    & 0.4765    & 0.8581 \\
    FixMatch \cite{sohn2020fixmatch}    & 0.9731    & 0.9706    & 0.9716    & 0.9962    & 0.6590    & 0.5830    & 0.5405    & 0.8862 \\ 
    FlexMatch \cite{zhang2021flexmatch}   & 0.9566    & 0.9691    & 0.9625    & 0.9975    & 0.6403    & 0.6755    & 0.6518    & 0.9249 \\
    \rowcolor{LightGreen}\textbf{ReFixMatch}   & 0.9779   & \textbf{0.9777}    & \textbf{0.9778}    & \textbf{0.9978}    & \textbf{0.8518}    & \textbf{0.7140}    & \textbf{0.6908}  & \textbf{0.9571}        \\ \midrule    \bottomrule
    \end{tabular}%
\end{sc}
}
\end{table*}

\subsubsection{Imbalance dataset problem}
\begin{figure}[!ht]
    \centering
    \subfloat[\footnotesize{FixMatch vs ReFixMatch}\label{fig:fdsvhn}]
    {\includegraphics[width=.49\linewidth]{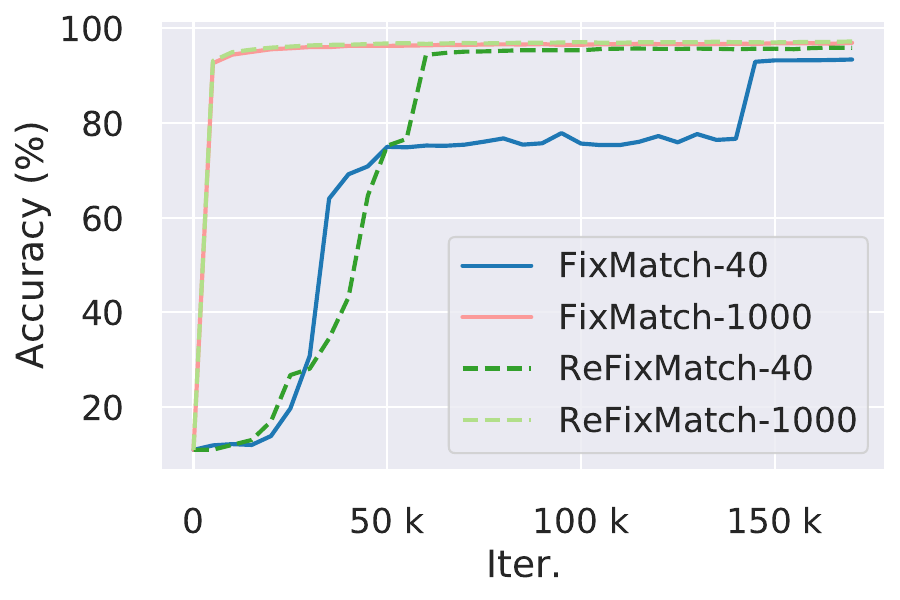}}%
    \subfloat[\footnotesize{FlexMatch vs ReFixMatch+CPL}\label{fig:fldsvhn}]
    {\includegraphics[width=.49\linewidth]{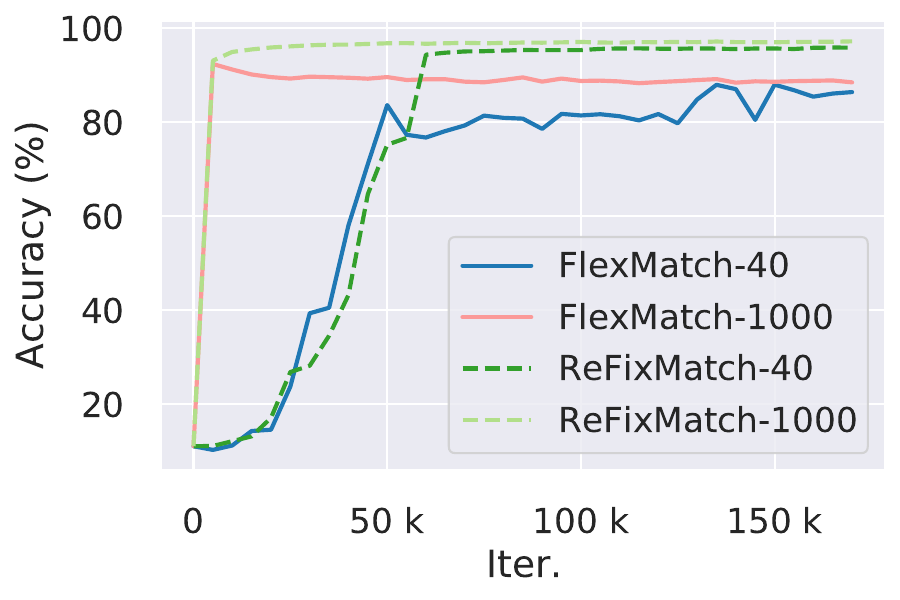}}
    \caption{Accuracy comparison of Figure \ref{fig:fdsvhn}: FixMatch vs ReFixMatch and Figure \ref{fig:fldsvhn}: FlexMatch vs ReFixMatch+CPL for first 150k iterations on SVHN dataset with 40-label and 1000-label split.}
\end{figure}

For example, when dealing with imbalanced datasets such as the SVHN and ImageNet datasets, ReFixMatch outperforms both FixMatch and FlexMatch.
FlexMatch fails on the SVHN dataset since CPL may yield low final thresholds for the tail classes, allowing noisy pseudo-labeled samples to be accepted and trained.
In contrast, ReFixMatch preserves the high fixed threshold of FixMatch, and the final results on the SVHN dataset are improved.
In addition, ReFixMatch outperforms both FixMatch and FlexMatch by a large margin without additional modules on the ImageNet.

\begin{figure}[!ht]
    \centering
    \subfloat[\footnotesize{Top-1 accuracy}\label{fig:imagenetacc}]
    {\includegraphics[width=.49\linewidth]{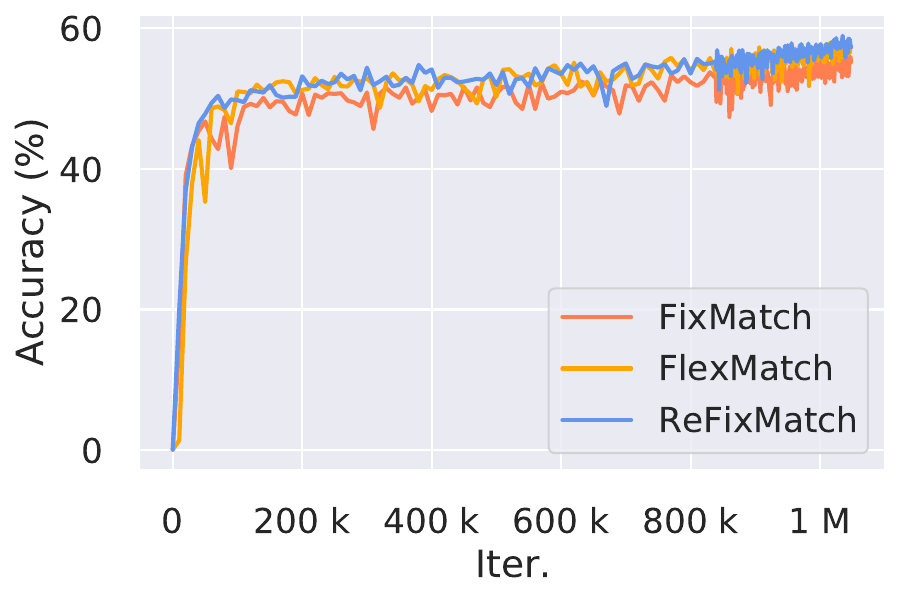}}%
    \subfloat[\footnotesize{Eval loss}\label{fig:imagenetloss}]
    {\includegraphics[width=.49\linewidth]{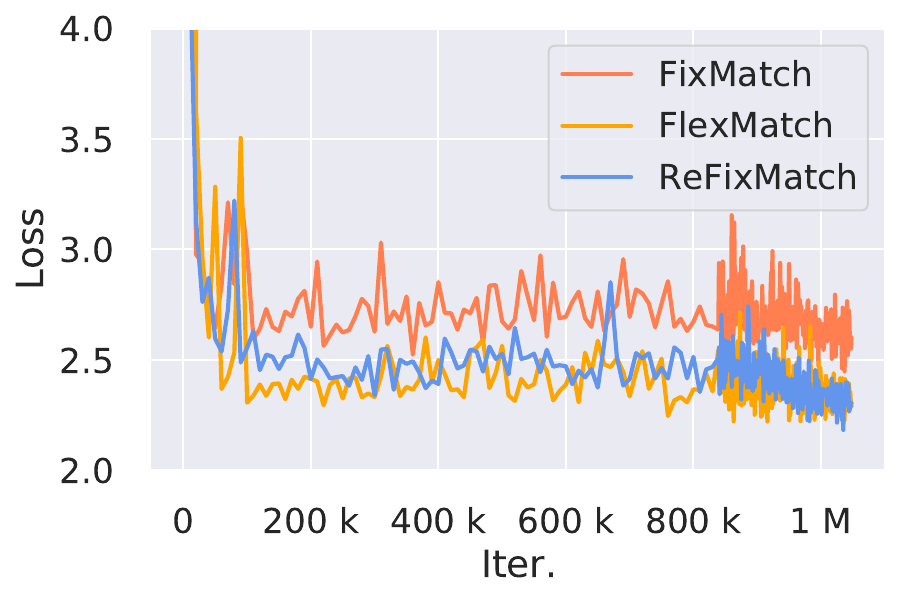}}
    \caption{Accuracy and loss comparison of FixMatch, FlexMatch, and ReFixMatch on ImageNet dataset.}
\end{figure}

\subsubsection{Long-tailed issue}
To further prove the effectiveness of ReFixMatch, we evaluate ReFixMatch on the imbalanced SSL setting.
We conduct experiments on CIFAR-10-LT, SVHN-LT, and CIFAR-100-LT with different imbalance ratios.
Following \cite{lee2021abc,wei2021crest,ren2020balanced}, we use WRN-28-2 as the backbone.
We consider long-tailed (LT) imbalance, where the number of data points exponentially decreases from the first class to the last, i.e., $N_k = N_1 \times \lambda^{-\frac{k-1}{L-1}}$, where $\lambda = \frac{N_1}{N_k}$.
For CIFAR-10, we set $\lambda=100, N_1=1000$, and $\beta=10\%, 20\%$, and $30\%$, respectively.
Similarly, we set $\lambda=100, N_1=1000$, and $\beta=20\%$ for SVHN.
And for CIFAR-100, we set $\lambda=20, N_1=200$, and $\beta=40$.
The results are recorded in Table \ref{table:results4-lt} with an average of three different runs.

\begin{table*}[ht!]
\centering
\caption{Overall accuracy under the long-tailed setting}
\label{table:results4-lt}
\resizebox{0.65\textwidth}{!}{
\begin{sc}
\begin{tabular}{lccccc}
\toprule    \midrule
                    & \multicolumn{3}{c}{CIFAR-10-LT}                               & SVHN-LT                   & CIFAR-100-LT \\ \midrule
\multirow{2}{*}{Algorithm}  & \multicolumn{3}{c}{$\lambda=100$}                     & $\lambda=100$             & $\lambda=20$ \\ 
                    & $\beta=10\%$      &   $\beta=20\%$  &  $\beta=30\%$           & $\beta=20\%$              & $\beta=40\%$ \\ \midrule  \midrule
Vanilla             & -                 & 55.3±1.30     & -                         &    77.0±0.67              &    40.1±1.15 \\
VAT \cite{Miyato2019VirtualAT}           & -                 & 55.3±0.88     & -                         &    81.3±0.47              &    40.4±0.34 \\
BALMS \cite{ren2020balanced}         & -                 & 70.7±0.59     & -                         &    87.6±0.53              &    50.2±0.54 \\ \midrule
FixMatch \cite{sohn2020fixmatch}       & 70.0±0.59         & 72.3±0.33     & 74.9±0.63                 &    88.0±0.30              &    51.0±0.20 \\
w/ CReST+PDA \cite{wei2021crest}  & 73.9±0.40         & 76.6±0.46     & 74.9±0.63                 &    89.1±0.69              &    51.6±0.29 \\
w/ DARP \cite{kim2020distribution}       & -                 & 73.7±0.98     & -                         &    88.6±0.19              &    51.4±0.37 \\
w/ DARP+cRT \cite{kim2020distribution}    & 74.6±0.98         & 78.1±0.895    & 77.6±0.73                 &    89.9±0.44              &    54.7±0.46 \\
w/ ABC \cite{lee2021abc}             & 77.2±1.60         & 81.1±0.82     & 81.5±0.29                 &    92.0±0.38              &    56.3±0.19 \\
\rowcolor{LightGreen}w/ ABC + \textbf{ReFixMatch} &  \textbf{85.4±0.01}                  & \textbf{81.3±0.75}     &      \textbf{82.1±0.25}                     &    \textbf{92.1±0.06}              &    \textbf{57.0±0.09} \\ \midrule   \bottomrule
\end{tabular}%
\end{sc}
}
\end{table*}

Surprisingly, ReFixMatch boosts the performance by a large margin when used with ABC \cite{lee2021abc}.
With an accuracy of 85.42\%, ReFixMatch outperforms ABC with an 8.2\% improvement when $\beta$ equals 10\%.

\subsection{Calibration of SSL}
\begin{figure*}[!ht]
    \centering
\resizebox{0.9\linewidth}{!}{
    \subfloat[\footnotesize{FixMatch}\label{fig:ece1}]
    {\includegraphics[width=.24\linewidth]{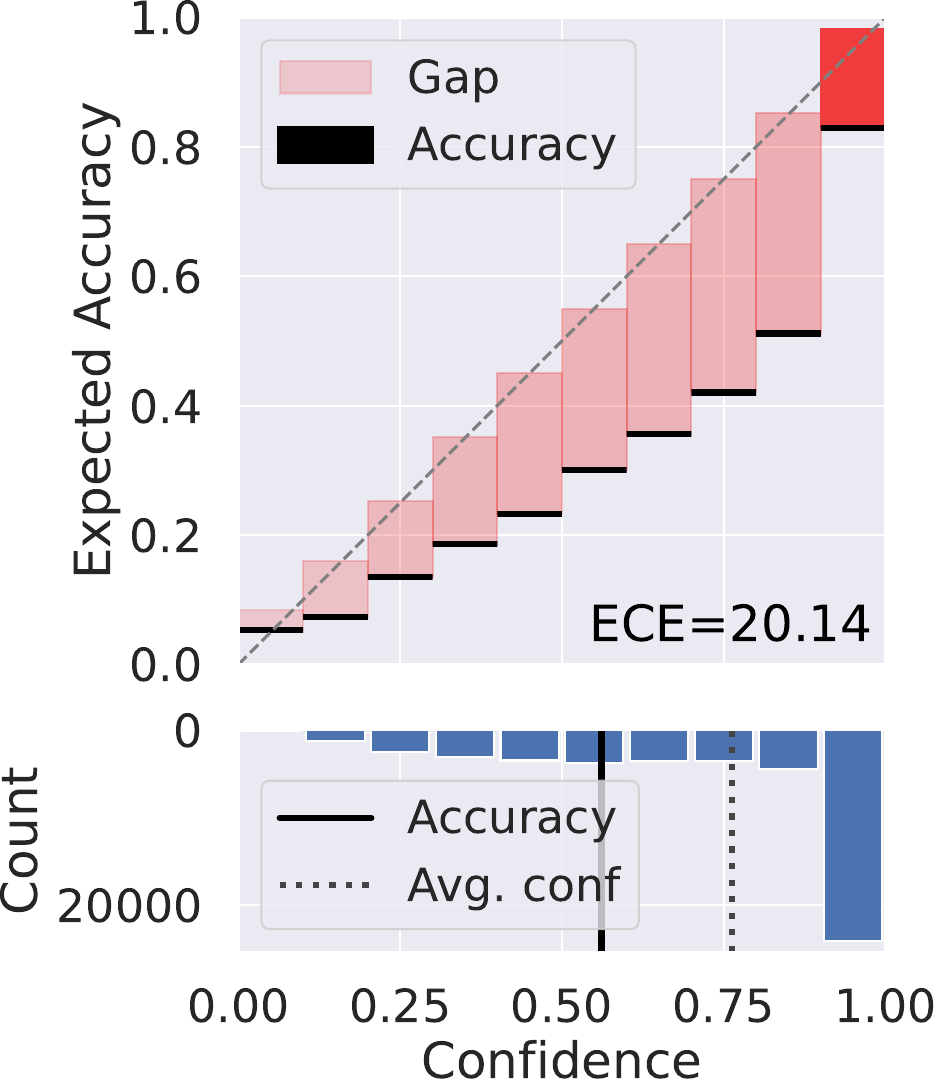}}%
    \subfloat[\footnotesize{FlexMatch}\label{fig:ece2}]
    {\includegraphics[width=.24\linewidth]{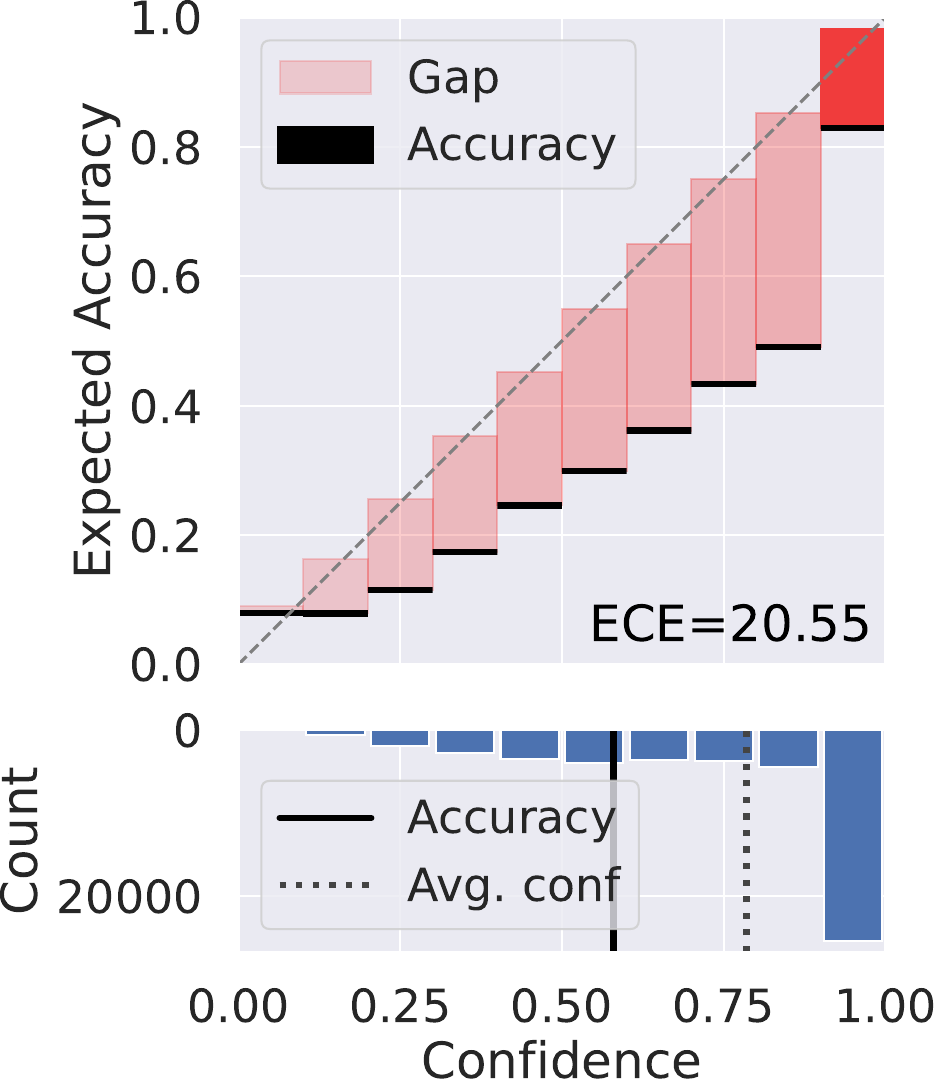}}%
    \subfloat[\footnotesize{ReFixMatch}\label{fig:ece3}]
    {\includegraphics[width=.24\linewidth]{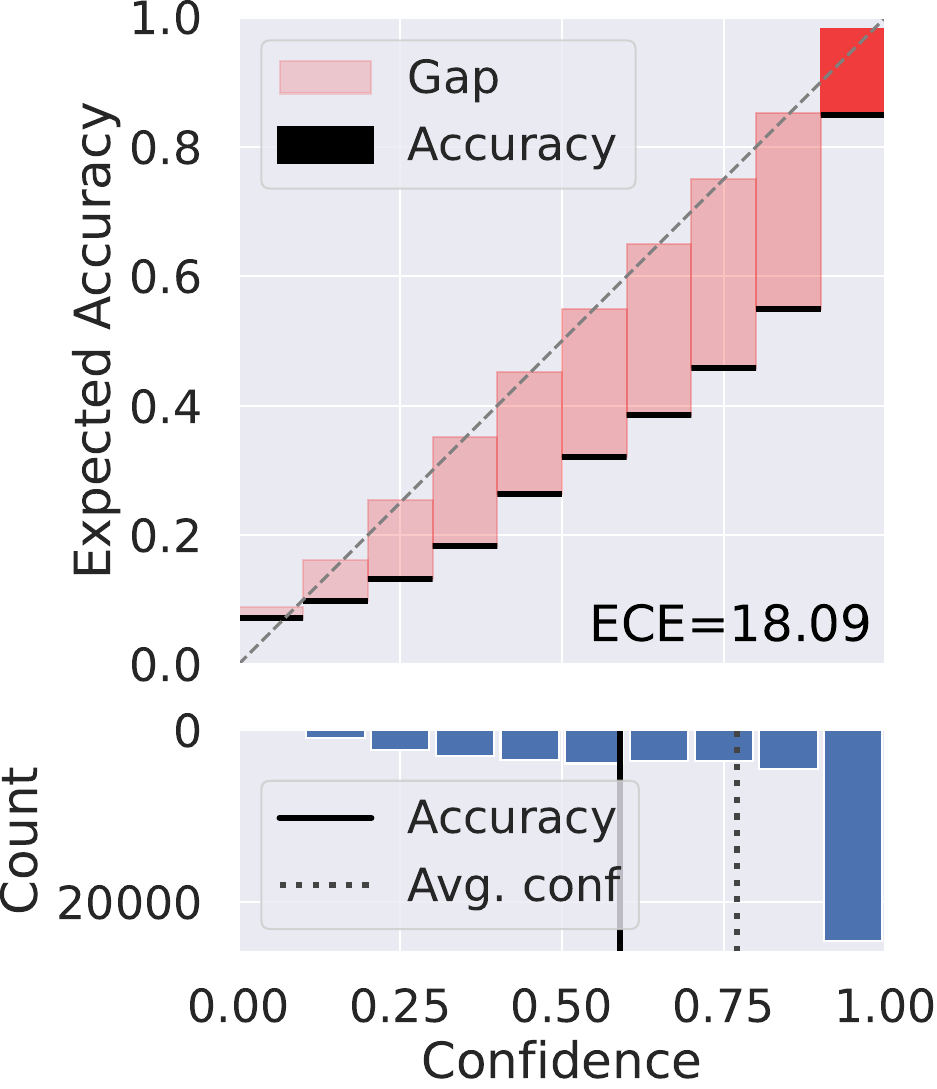}}%
    \subfloat[\footnotesize{ReFixMatch+CPL}\label{fig:ece4}]
    {\includegraphics[width=.24\linewidth]{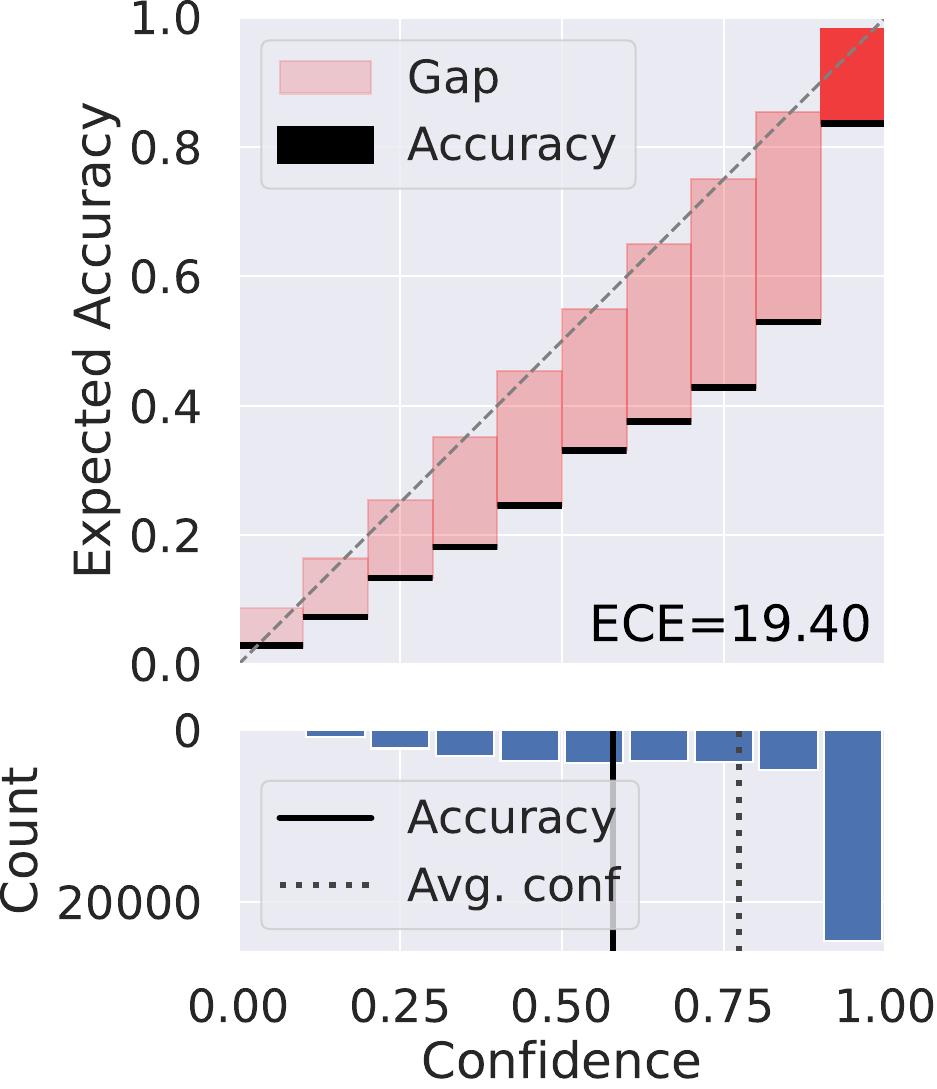}}%
    }
    \caption{Reliability diagrams (top) and confidence histograms (bottom) for ImageNet dataset.}
    \label{fig:ece}
\end{figure*}

\cite{chen2022semi} suggests addressing confirmation bias from the calibration perspective.
We measure the calibration of FixMatch, FlexMatch, ReFixMatch, and ReFixMatch+CPL trained on the ImageNet dataset with 100k labels \footnote{https://github.com/hollance/reliability-diagrams}.
Several common calibration indicators are used: Expected Calibration Error (ECE), confidence histogram, and reliability diagram.
As shown in Fig. \ref{fig:ece}, even though FlexMatch has higher accuracy than FixMatch, its ECE value of 20.55 is larger than that of FixMatch, at 20.14, indicating poorer probability estimation.
On the other hand, ReFixMatch achieves both higher accuracy and a lower ECE value of 18.09, which proves that it can reduce the confirmation bias and produce a better calibrated model.
Furthermore, despite having a lower performance than FlexMatch, ReFixMatch+CPL still achieves an ECE value of 19.40.

\section{Related Work}
In SSL, self-training is extensively used \cite{mclachlan1975iterative,scudder1965probability}. 
The model's output probabilities are treated as "soft" labels for unlabeled data. 
Pseudo-labeling is a self-training variation that converts the probability to "hard" labels \cite{Lee2013PseudoLabelT}. 
To alleviate the confirmation bias problem, pseudo-labeling is used together with confidence-based thresholding, which keeps unlabeled samples only when predictions are sufficiently confident \cite{Rosenberg2005SemiSupervisedSO,Xie2020UnsupervisedDA,sohn2020fixmatch,zhang2021flexmatch}.
Consistency regularization is used to make predictions on perturbed versions of unlabeled data match the pseudo-label \cite{Bachman2014LearningWP,Laine2017TemporalEF,Sajjadi2016RegularizationWS}.
There are many techniques to generate perturbed versions of unlabeled data such as data augmentation \cite{french2017self}, stochastic regularization \cite{Laine2017TemporalEF,sajjadi2016regularization}, and adversarial perturbations \cite{Miyato2019VirtualAT}.

FixMatch \cite{sohn2020fixmatch} presents a hybrid approach for SSL that combines pseudo-labeling and consistency regularization.
The qualified pseudo-labeling in FixMatch creates a sharpening-like effect that promotes the ability of the model to give high-confidence predictions.
FlexMatch proposes a Curriculum Pseudo Labeling (CPL) approach, which allows standard SSL to train with a dynamic threshold for each class.
CPL is a dynamic thresholding strategy since it dynamically adjusts the threshold for each class after each iteration, allowing better performance for each class.

\cite{li2021comatch} propose CoMatch, which combines the ideas of consistency regularization and contrastive learning, in which the target similarity of two instances is measured by the similarity of two class probability distributions, and it achieves the current state-of-the-art semi-supervised learning performance.
However, the hyperparameters are extremely sensitive, and the optimal temperature and threshold vary for different datasets and settings.

SimMatch \cite{zheng2022simmatch} proposes a novel semi-supervised learning framework that simultaneously considers semantic similarity and instance similarity.
It shows that by considering consistency regularization on both the semantic level and instance level, SimMatch improves its performance and achieves state-of-the-art semi-supervised learning.

% Unsupervised Semantic Aggregation and Deformable Template Matching (USADTM) is a method proposed by \cite{han2020unsupervised} to improve classification performance with minimally labeled data and thus lower the cost of data annotation.
% \cite{gong2021alphamatch} proposes AlphaMatch, an efficient SSL approach that utilizes data augmentations by effectively enforcing label consistency between data points and the augmented data obtained from them.

\section{Conclusions}
In this paper, we present ReFixMatch, a new semi-supervised learning pipeline that improves upon the conventional FixMatch algorithm by utilizing both high-confidence and low-confidence predictions. 
% We show that training FixMatch with only low-confidence predictions can still produce high performance.
% Specifically, ReFixMatch redefines conventional FixMatch with an extra loss term for low-confidence predictions.
Despite its simplicity, ReFixMatch can significantly improve the generalization of the model and boost performance without any additional computational overheads. 
% The results on ImageNet demonstrate that ReFixMatch achieves state-of-the-art performance for semi-supervised learning, especially on imbalanced datasets.
ReFixMatch outperforms the conventional state-of-the-art methods by a large margin across datasets without introducing additional modules or computational overheads.

\section{Acknowledgement}
This work was supported by the Institute of Information and Communications Technology Planning and Evaluation (IITP) Grant funded by the Korean Government through the Ministry of Science and ICT (MSIT) under Grant 2021-0-00106 and Grant 2022-0-00971.
% APPENDIX
%%%%%%%%%%%%%%%%%%%%%%%%%%%%%%%%%%%%%%%%%%%%%%%%%%%%%%%%%%%%%%%%%%%%%%%%%%%%%%%
%%%%%%%%%%%%%%%%%%%%%%%%%%%%%%%%%%%%%%%%%%%%%%%%%%%%%%%%%%%%%%%%%%%%%%%%%%%%%%%
\newpage
\appendix
\section*{Hyperparameter setting}
\label{sec:app}
We show the detailed training hyperparameter settings for each method in Table \ref{table:hyperparameter1}.
We also report the detailed hyperparameter settings with a specific model for each dataset in Table \ref{table:hyperparameter2}.

\begin{table*}[ht!]
\centering
    \caption{Training hyperparameters}
    \label{table:hyperparameter1}
% \resizebox{\textwidth}{!}{
\begin{sc}
    \begin{tabular}{@{}l|cccc@{}}
    \toprule    \midrule
    Algorithm   & UDA   & ReFixMatch  & FixMatch (FlexMatch)          \\ \midrule   \midrule
    Unlabeled Data to Labeled Data Ratio   & \multirow{2}{*}{7}     & \multirow{2}{*}{7}    & \multirow{2}{*}{7} \\
    (CIFAR-10/100, STL-10, SVHN)   &       &        &       \\ \midrule
    Unlabeled Data to Labeled Data Ratio     & \multirow{2}{*}{-}         & \multirow{2}{*}{1}         & \multirow{2}{*}{1} \\ 
    (ImageNet)          &           &           & \\ \midrule
    Pre-defined Threshold                               & \multirow{2}{*}{0.8}  & \multirow{2}{*}{0.95}       & \multirow{2}{*}{0.95}                             \\
    (CIFAR-10/100, STL-10, SVHN)                        &           &           & \\ \midrule
    Pre-defined Threshold (ImageNet)                    & -         & 0.7       & 0.7 \\ \midrule 
    Temperature                                         & 0.4       & 0.5       & - \\ \midrule \bottomrule
    \end{tabular}%
\end{sc}
% }
\end{table*}

\begin{table*}[ht!]
\centering
    \caption{Dataset-wise hyperparameters}
    \label{table:hyperparameter2}
\begin{sc}
    \begin{tabular}{@{}l|ccccc}
    \toprule    \midrule
    Dataset         & CIFAR-10      & CIFAR-100     & STL-10    & SVHN      & ImageNet  \\ \midrule \midrule
    Model           & WRN-28-2      & WRN-28-8      & WRN-37-2  & WRN-28-2  & ResNet-50 \\ \midrule
    Weight Decay    & 5e-4          & 1e-3          & 5e-4      & 5e-4      & 3e-4      \\ \midrule
    Batch Size      &  \multicolumn{4}{c}{64}                               & 128       \\ \midrule
    Learning Rate   & \multicolumn{5}{c}{0.03}                                          \\ \midrule
    SGD Momentum    & \multicolumn{5}{c}{0.9}                                           \\ \midrule
    EMA Momentum    & \multicolumn{5}{c}{0.999}                                         \\ \midrule
    Unsupervised Loss Weight & \multicolumn{5}{c}{1}                                    \\ \midrule \bottomrule
    \end{tabular}%
\end{sc}
\end{table*}

\section*{Detailed results}
\label{app:details}
Following the suggestion from \cite{sohn2020fixmatch}, we also report the median error rates of the last 20 checkpoints in Table \ref{table:mer}.
The results show that our proposed ReFixMatch improves performance and surpasses previous methods by a large margin.
Furthermore, the results also show that the model trained using ReFixMatch keeps improving until the end of the training process, while FlexMatch is overfit to the data.

\begin{table*}[ht!]
\centering
    \caption{Mean error rates of last 20 checkpoints of all methods. 
    There are 1000 iterations between every two checkpoints}
    \label{table:mer}
\resizebox{\textwidth}{!}{
\begin{sc}
    \begin{tabular}{@{}l|ccc|ccc|ccc|cc}
    \toprule    \midrule
Dataset &   \multicolumn{3}{c|}{CIFAR-10}    & \multicolumn{3}{c|}{CIFAR-100} & \multicolumn{3}{c|}{SVHN}    &   \multicolumn{2}{c}{STL-10}\\
\midrule
\# Label &   40  &   250 &   4000    &   400 &   2500    &   10000   &   40  &   250 &   1000   &   40  &   1000\\
\midrule    \midrule
$\Pi$ Model   &   78.78\tiny{$\pm$}2.24  &   55.79\tiny{$\pm$}2.61  &   13.63\tiny{$\pm$}0.60  &   89.27\tiny{$\pm$}0.73  &   60.58\tiny{$\pm$}0.66 &   38.49\tiny{$\pm$}0.09  &   76.23\tiny{$\pm$}4.60  &   18.44\tiny{$\pm$}2.79  &   7.77\tiny{$\pm$}0.03   &   77.80\tiny{$\pm$}0.63  &   35.63\tiny{$\pm$}0.25\\
Pseudo Label    &   77.42\tiny{$\pm$}1.19  &   48.33\tiny{$\pm$}2.43  &   15.64\tiny{$\pm$}0.29  &   90.01\tiny{$\pm$}0.21  &  58.38\tiny{$\pm$}0.42  &   37.64\tiny{$\pm$}0.16  &   69.05\tiny{$\pm$}6.77  &   16.76\tiny{$\pm$}1.02  &   9.99\tiny{$\pm$}0.35   &   76.44\tiny{$\pm$}0.67  &  33.57\tiny{$\pm$}0.40\\
VAT &   81.90\tiny{$\pm$}2.39  &   42.43\tiny{$\pm$}1.86  &   10.83\tiny{$\pm$}0.07  &   89.28\tiny{$\pm$}1.71  &   47.44\tiny{$\pm$}0.68  &   32.66\tiny{$\pm$}0.33  &   80.19\tiny{$\pm$}4.08  &   4.54\tiny{$\pm$}0.12   &   4.31\tiny{$\pm$}0.20   &   78.34\tiny{$\pm$}1.24  &   48.36\tiny{$\pm$}0.29\\
Mean Teacher    &   77.96\tiny{$\pm$}2.63  &   42.47\tiny{$\pm$}3.79  &   8.49\tiny{$\pm$}0.21   &   81.58\tiny{$\pm$}1.51  &  45.61\tiny{$\pm$}1.12  &   32.38\tiny{$\pm$}0.12  &   47.12\tiny{$\pm$}2.96  &   3.56\tiny{$\pm$}0.04   &   3.38\tiny{$\pm$}0.03   &   76.04\tiny{$\pm$}2.94  &  38.94\tiny{$\pm$}1.14\\
UDA &   10.96\tiny{$\pm$}3.68  &   5.46\tiny{$\pm$}0.07   &   4.60\tiny{$\pm$}0.05   &   51.97\tiny{$\pm$}1.38  &   29.92\tiny{$\pm$}0.35  &   23.64\tiny{$\pm$}0.33  &   5.31\tiny{$\pm$}4.39   &   \textbf{2.01\tiny{$\pm$}0.03}   &   1.97\tiny{$\pm$}0.04   &   41.11\tiny{$\pm$}5.21  &   8.00\tiny{$\pm$}0.58\\
FixMatch    &   7.99\tiny{$\pm$}0.59   &   \textbf{5.12\tiny{$\pm$}0.03}   &   4.46\tiny{$\pm$}0.11   &   48.95\tiny{$\pm$}1.19  &   29.19\tiny{$\pm$}0.25 &   23.06\tiny{$\pm$}0.12  &   3.92\tiny{$\pm$}1.18   &   2.09\tiny{$\pm$}0.03   &   2.06\tiny{$\pm$}0.01   &   44.70\tiny{$\pm$}6.58  &   7.38\tiny{$\pm$}0.26\\
Dash    &   11.02\tiny{$\pm$}4.05  &   5.43\tiny{$\pm$}0.20   &   4.68\tiny{$\pm$}0.07   &   47.88\tiny{$\pm$}1.31  &   28.62\tiny{$\pm$}0.41  &   22.92\tiny{$\pm$}0.15  &   2.28\tiny{$\pm$}0.18   &   2.12\tiny{$\pm$}0.04   &   2.07\tiny{$\pm$}0.01   &   41.21\tiny{$\pm$}5.25  &   7.52\tiny{$\pm$}0.81\\
MPL &   9.65\tiny{$\pm$}3.02   &   6.08\tiny{$\pm$}0.48   &   4.76\tiny{$\pm$}0.06   &   48.45\tiny{$\pm$}1.61  &   28.41\tiny{$\pm$}0.14  &   \textbf{22.25\tiny{$\pm$}0.18}  &   14.74\tiny{$\pm$}14.69 &   2.41\tiny{$\pm$}0.04   &   2.39\tiny{$\pm$}0.01   &   41.49\tiny{$\pm$}3.90  &   7.05\tiny{$\pm$}0.51\\
FlexMatch   &   5.19\tiny{$\pm$}0.05   &   5.33\tiny{$\pm$}0.12   &   4.47\tiny{$\pm$}0.09   &   45.91\tiny{$\pm$}1.76  &   28.11\tiny{$\pm$}0.20 &   23.04\tiny{$\pm$}0.28  &   20.81\tiny{$\pm$}5.26  &   17.32\tiny{$\pm$}2.07  &   12.90\tiny{$\pm$}2.68  &   44.69\tiny{$\pm$}7.49  &   \textbf{6.15\tiny{$\pm$}0.25}\\
\rowcolor{LightGreen}\textbf{ReFixMatch}   &   \textbf{5.03\tiny{$\pm$}0.11}   &   5.16\tiny{$\pm$}0.10   &   \textbf{4.43\tiny{$\pm$}0.02}   &   \textbf{44.52\tiny{$\pm$}1.01}  &   \textbf{27.95\tiny{$\pm$}0.22} &   23.01\tiny{$\pm$}0.18  &   \textbf{2.20\tiny{$\pm$}0.34}  &   2.03\tiny{$\pm$}0.03  &   \textbf{2.01\tiny{$\pm$}0.01}  &   \textbf{40.21\tiny{$\pm$}6.11}  &   6.54\tiny{$\pm$}0.26\\ \midrule   \bottomrule
    \end{tabular}%
\end{sc}
}
\end{table*}

\section*{Qualitative Analysis}
\begin{figure}[!ht]
    \centering
    \subfloat[FixMatch\label{fig:tsne-fix}]
    {\includegraphics[width=.33\linewidth]{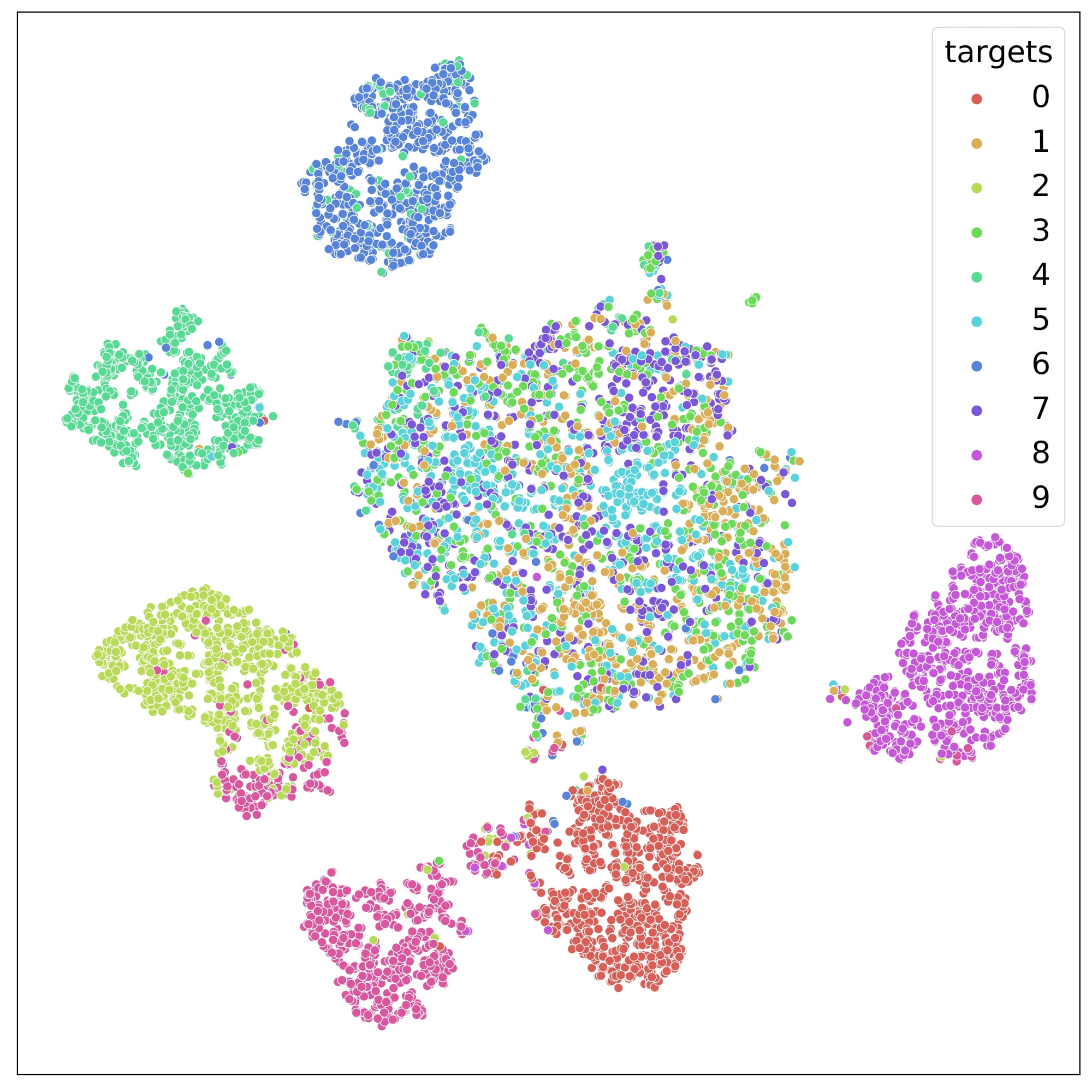}}%
    \subfloat[FlexMatch\label{fig:tsne-flex}]
    {\includegraphics[width=.33\linewidth]{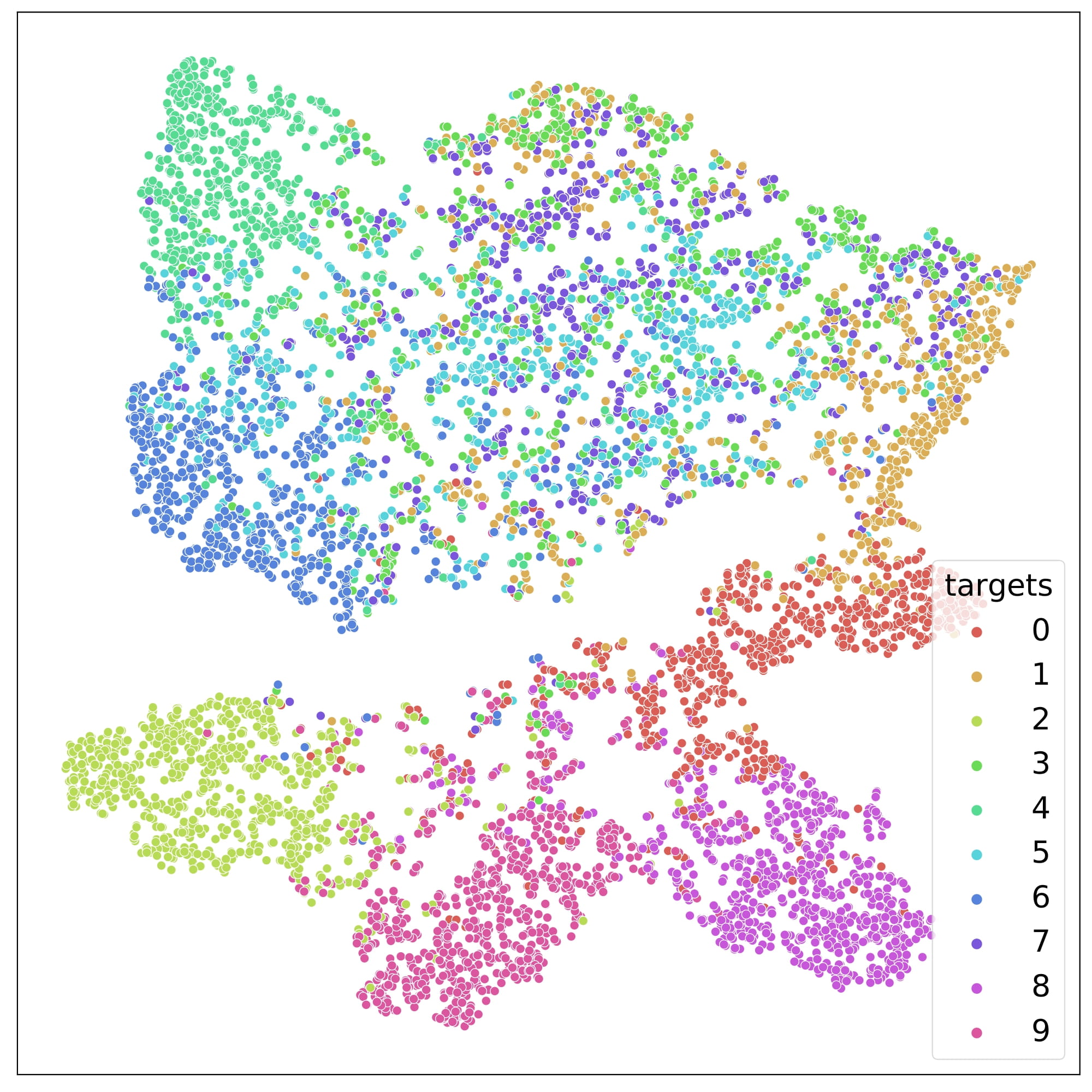}}%
    \subfloat[ReFixMatch\label{fig:tsne-refix}]
    {\includegraphics[width=.33\linewidth]{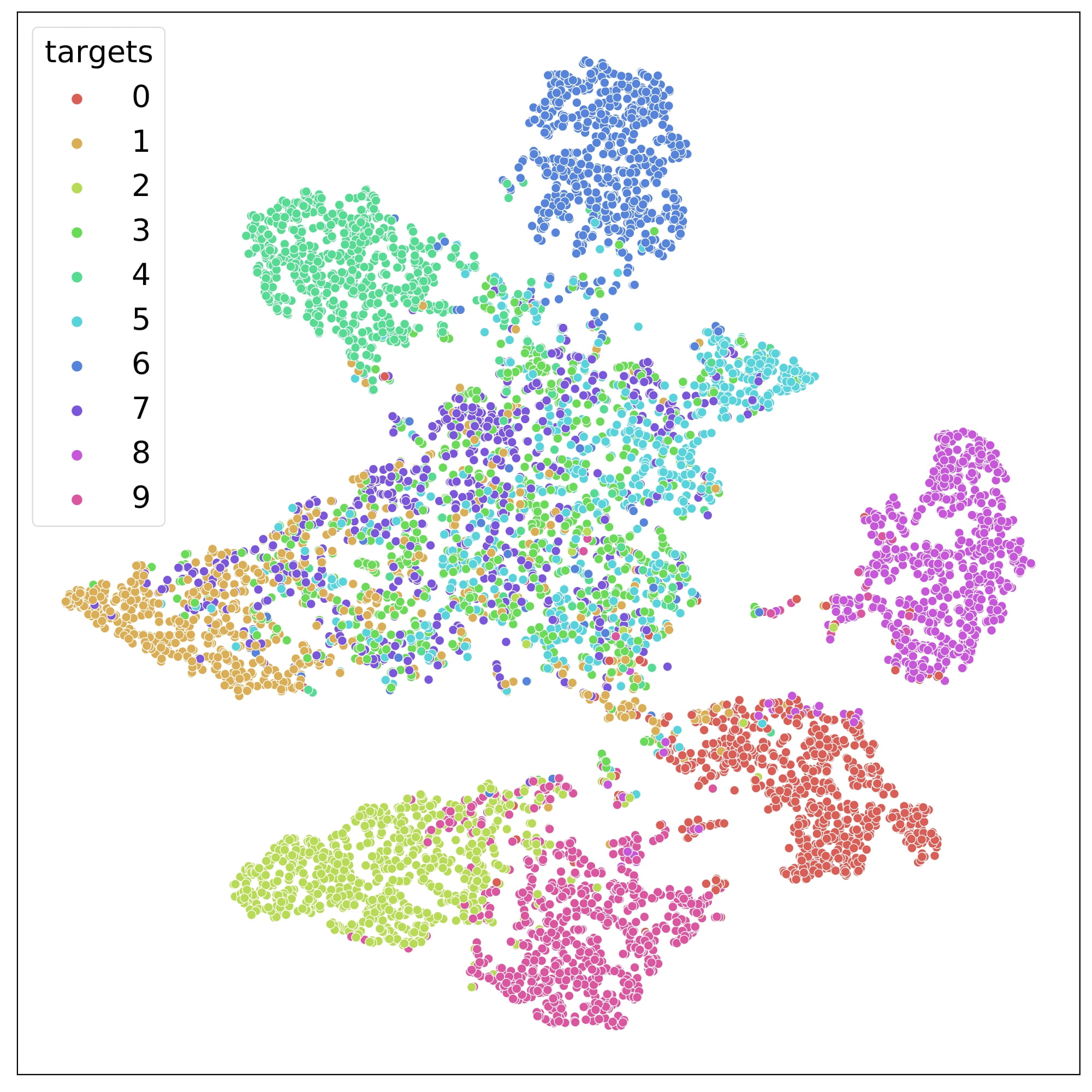}}%
    \caption{T-SNE visualization on STL-10 dataset with 40 labels.}
\end{figure}

We present the T-SNE visualization of features on STL-10 test dataset with 40-label split in Figure \ref{fig:tsne-fix},\ref{fig:tsne-flex},\ref{fig:tsne-refix}.
The visualization is using trained model from FixMatch, FlexMatch and ReFixMatch.

\begin{figure}[!ht]
    \centering
    \subfloat[FixMatch\label{fig:tsne-fix-svhn}]
    {\includegraphics[width=.33\linewidth]{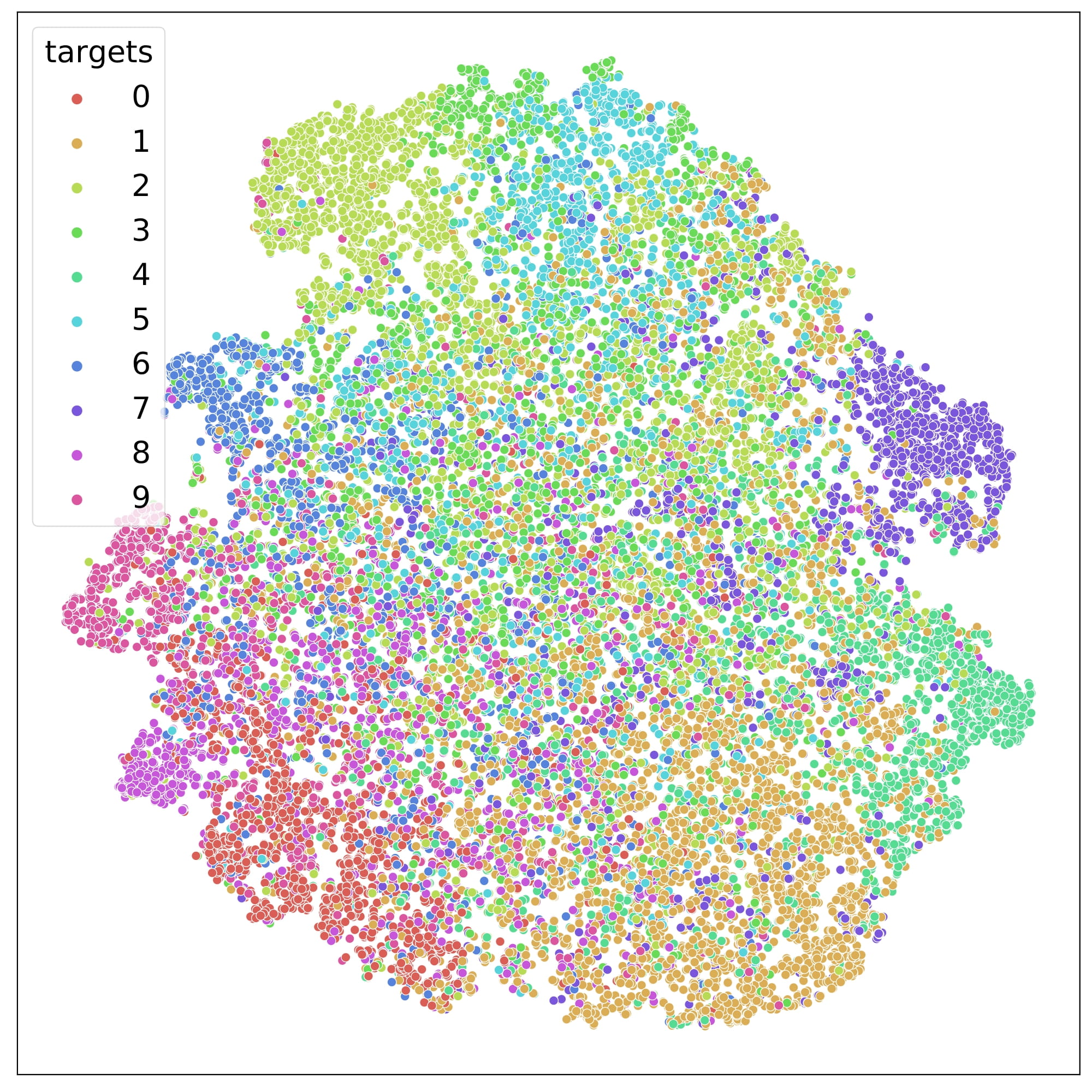}}%
    \subfloat[FlexMatch\label{fig:tsne-flex-svhn}]
    {\includegraphics[width=.33\linewidth]{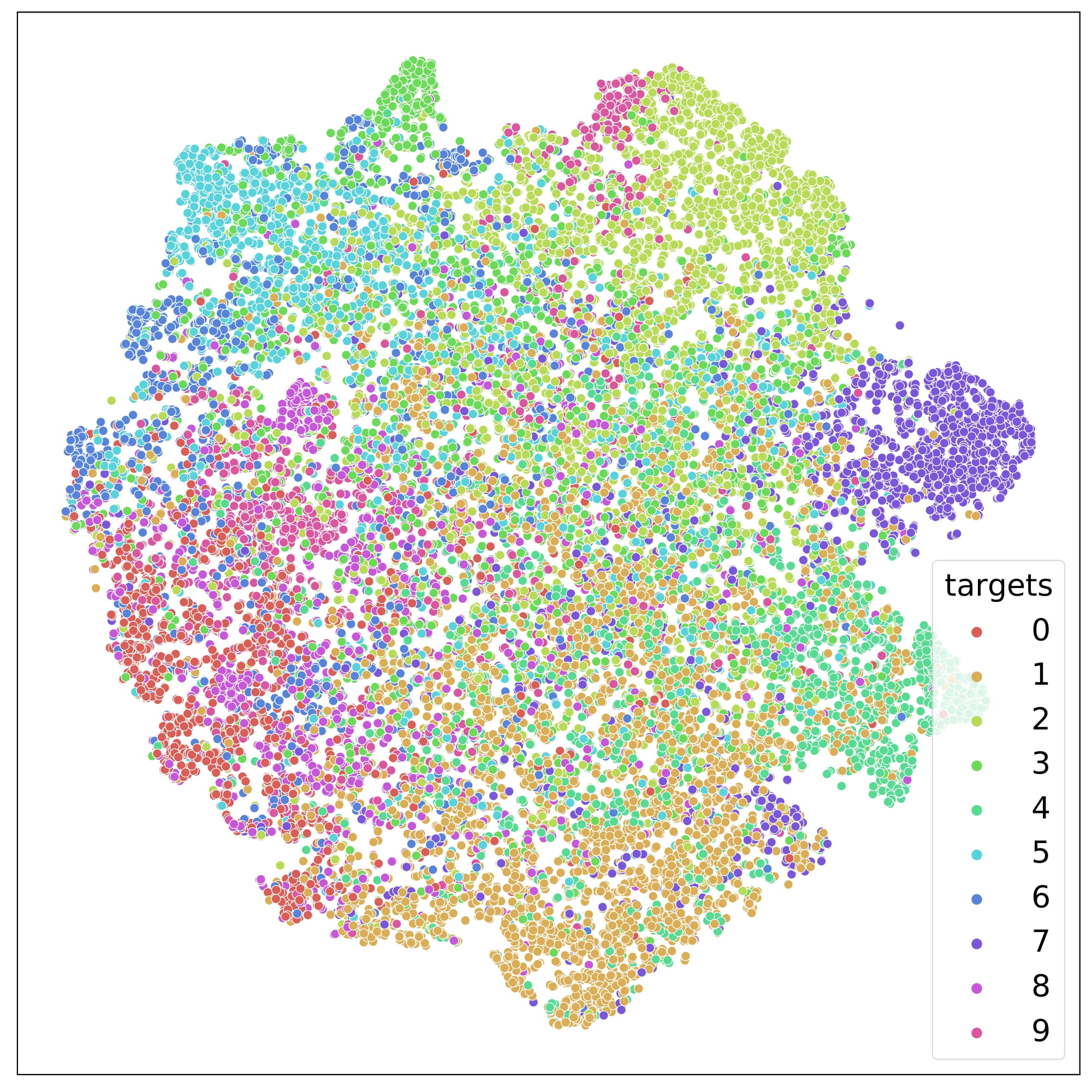}}%
    \subfloat[ReFixMatch\label{fig:tsne-refix-svhn}]
    {\includegraphics[width=.33\linewidth]{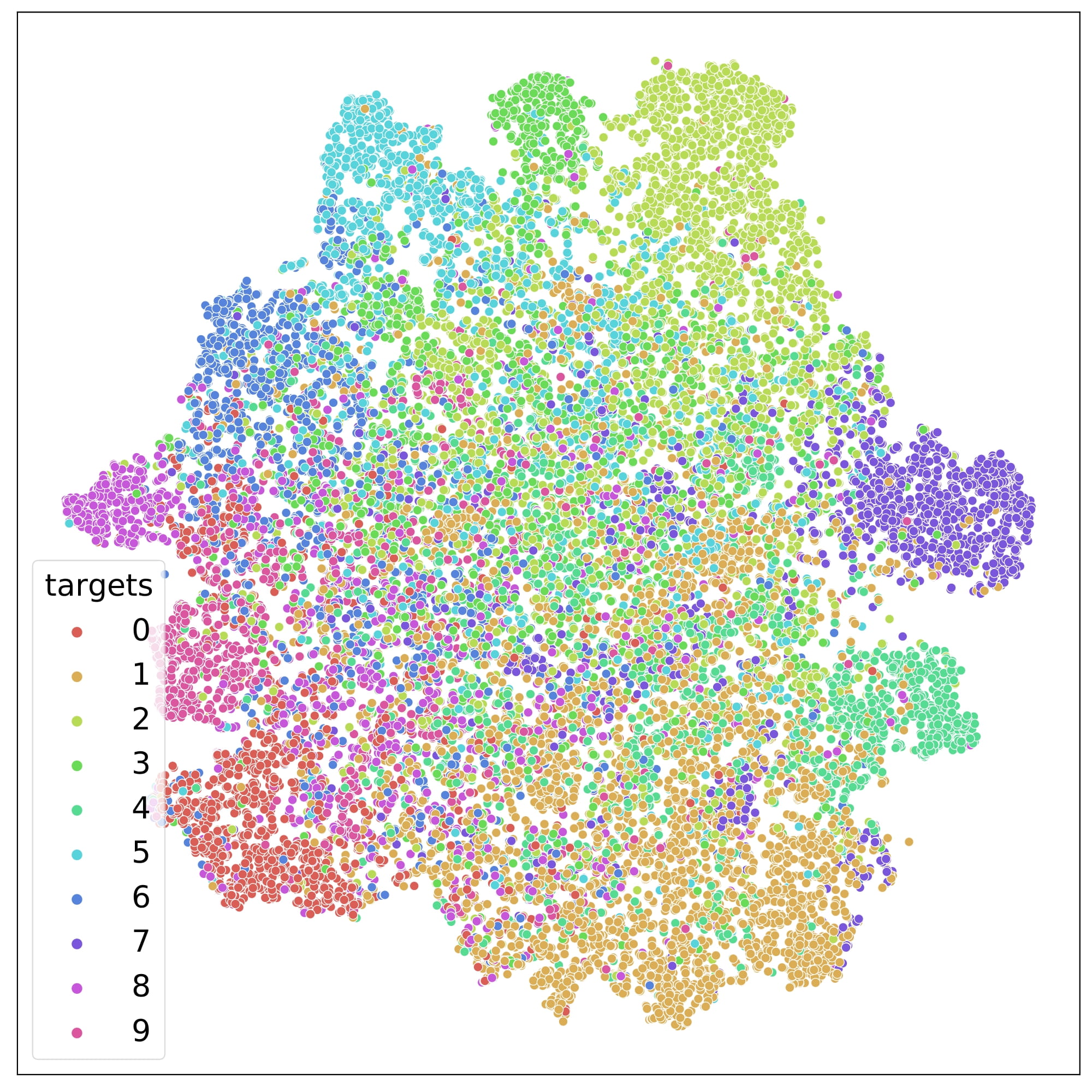}}%
    \caption{T-SNE visualization on SVHN dataset with 40 labels.}
\end{figure}

\begin{figure}[!ht]
    \centering
    \subfloat[FixMatch\label{fig:tsne-fix-cifar}]
    {\includegraphics[width=.33\linewidth]{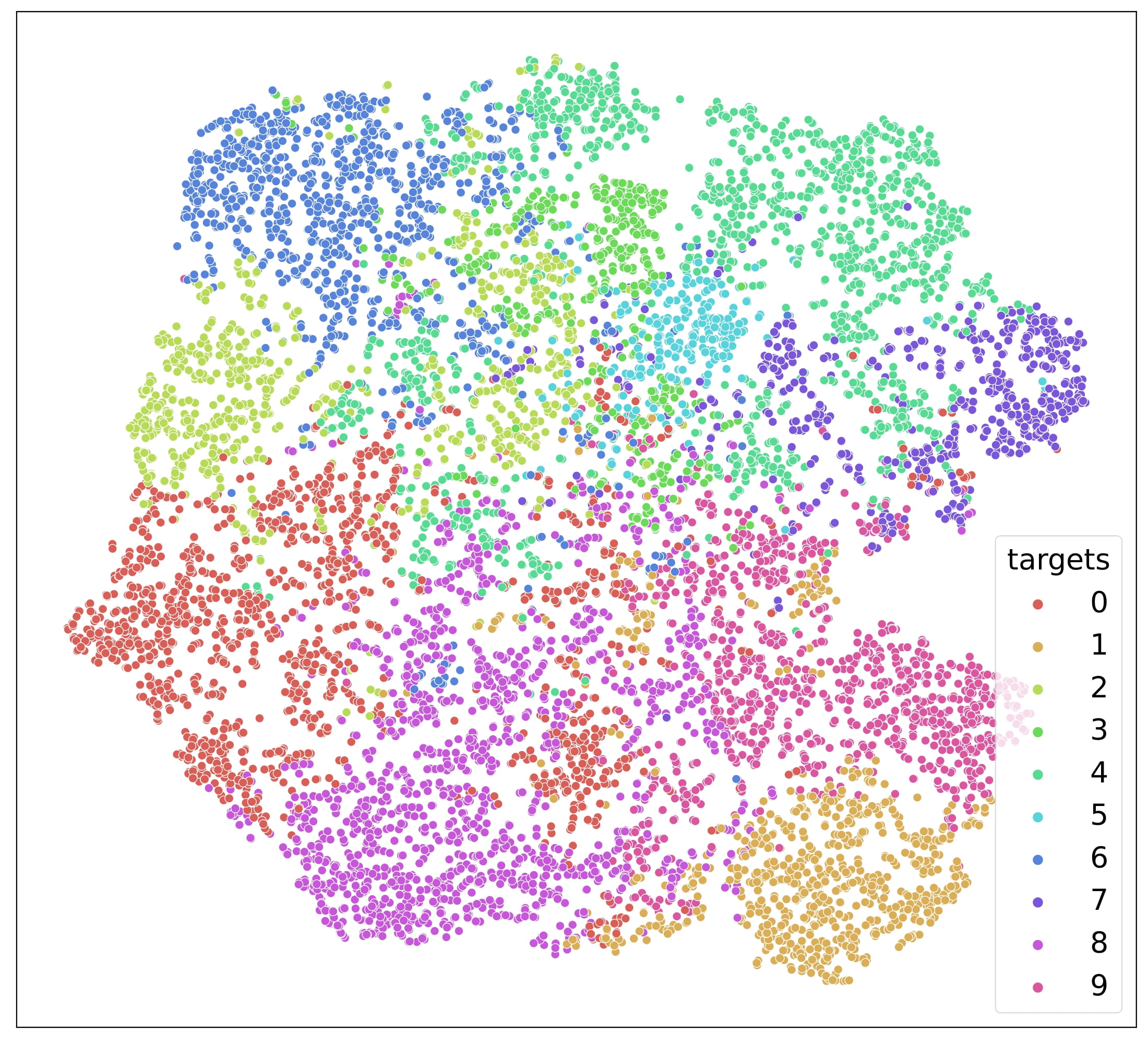}}%
    \subfloat[FlexMatch\label{fig:tsne-flex-cifar}]
    {\includegraphics[width=.33\linewidth]{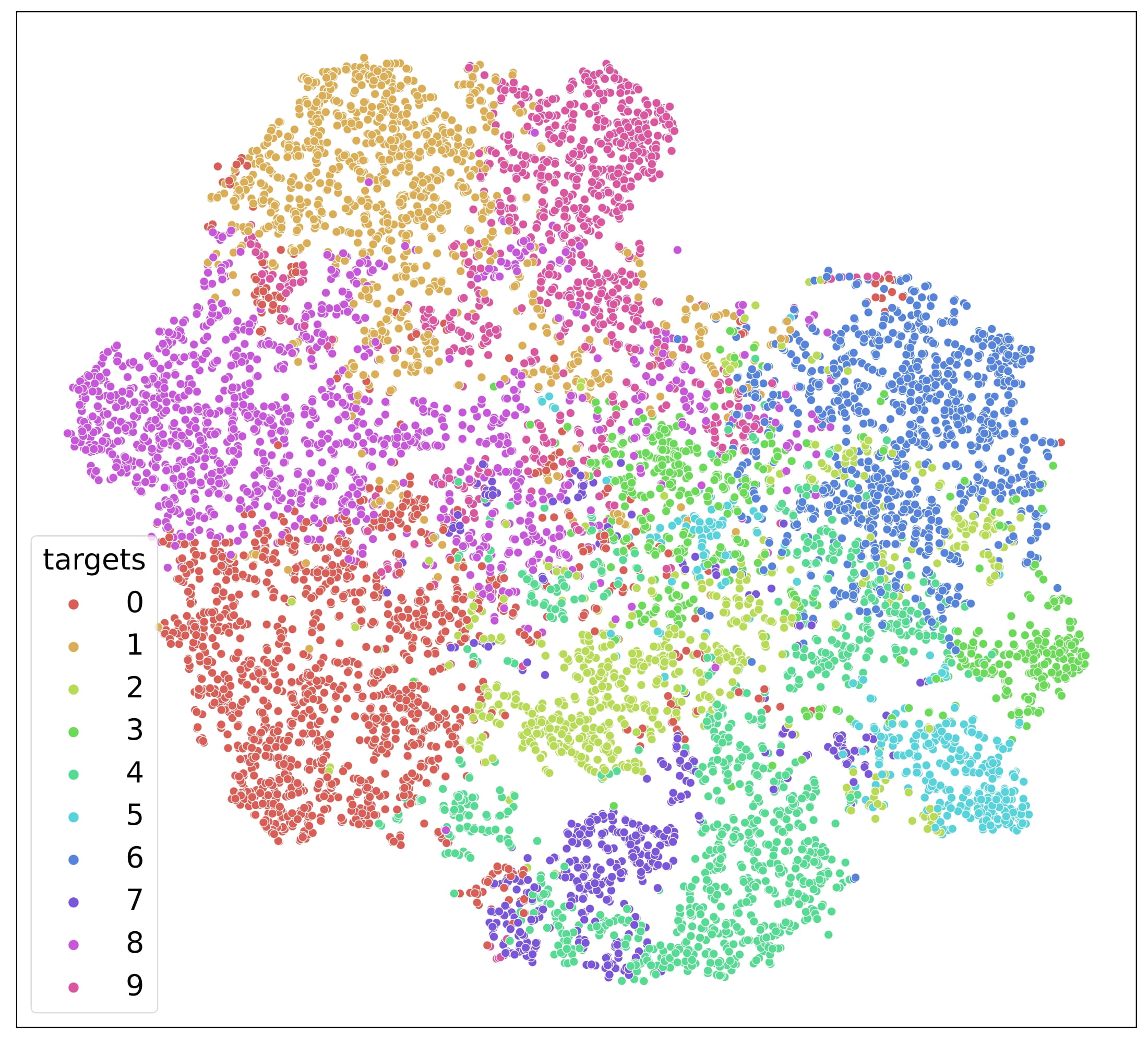}}%
    \subfloat[ReFixMatch\label{fig:tsne-refix-cifar}]
    {\includegraphics[width=.33\linewidth]{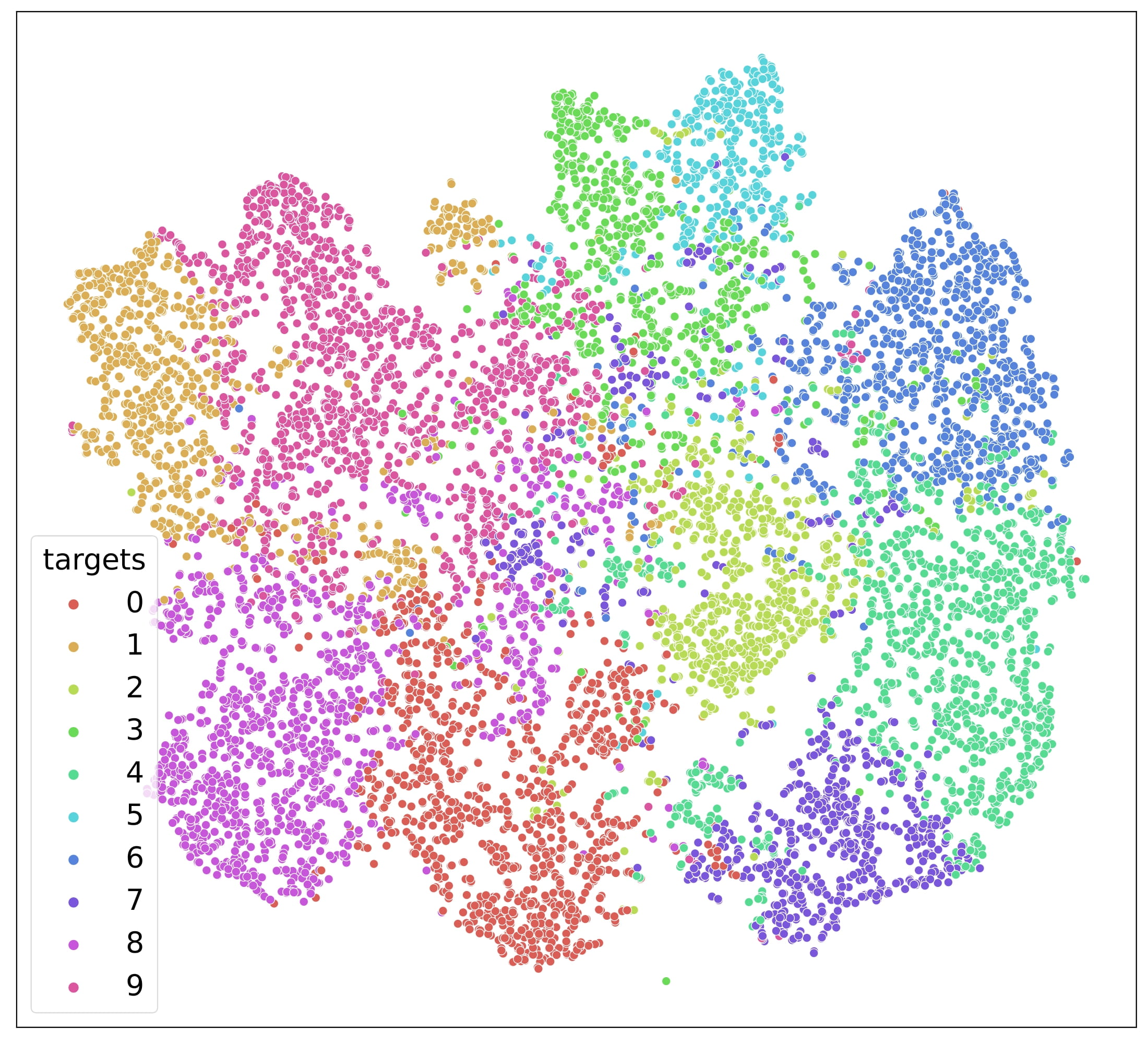}}%
    \caption{T-SNE visualization on CIFAR-10 dataset with 40 labels.}
\end{figure}

Figures \ref{fig:tsne-fix-svhn}, \ref{fig:tsne-flex-svhn}, and \ref{fig:tsne-refix-cifar} show the T-SNE visualization of features on the SVHN test dataset and the CIFAR-10 test dataset with a 40-label split.

As we can see, ReFixMatch produces a much clearer boundary for each class.
This clearly shows that ReFixMatch improves the generalization of the model.
In addition, we could see that although FlexMatch gives high performance, its border for class separation is not clear, this is due to the use of low threshold.

% \clearpage
\section*{ImageNet detailed results}
Table \ref{table:results3} shows the detailed results from Table \ref{table:results2}.
ReFixMatch without using self-supervised pre-trained weights outperforms previous methods such as CoMatch \cite{li2021comatch} and SimMatch \cite{zheng2022simmatch}.
ReFixMatch achieves 75.2\% of top-1 accuracy with the same training duration ($\sim400$ epochs) and has fewer parameters of 25.6M during training compared to 30.0M for FixMatch-EMAN, CoMatch, and SimMatch.

\begin{table*}[!ht]
    \centering
    \caption{Accuracy results on ImageNet with \textbf{10\%} labeled examples using \cite{li2021comatch} and \cite{zheng2022simmatch} source code.}
    \label{table:results3}
\begin{sc}
    \begin{tabular}{ll|cccl}
    \toprule    \midrule
    Self-supervised &   \multirow{2}{*}{Method}         & \multirow{2}{*}{Top-1}          & \multirow{2}{*}{Top-5}          & Params &   Epochs             \\
    Pre-training    &   &           &           & (train/test)      &  \\
    \midrule    \midrule
    None    &   FixMatch                                    & 71.5              & 89.1         & 25.6M/25.6M  &   $\sim300$              \\ 
    MoCo-EMAN\cite{cai2021exponential} &   FixMatch-EMAN\cite{cai2021exponential}     & 74.0              & 90.9         & 30.0M/25.6M  &   $\sim1100$               \\
    None &   CoMatch\cite{li2021comatch}                & 73.6              & 91.6         & 30.0M/25.6M  &   $\sim400$                \\ 
    MoCo V2\cite{chen2020improved} &   CoMatch\cite{li2021comatch}                & 73.7              & 91.4         & 30.0M/25.6M  &   $\sim1200$                \\ 
    None    &   SimMatch\cite{zheng2022simmatch}           & 74.4              & 91.6         & 30.0M/25.6M  &   $\sim400$             \\ 
    \midrule
    \rowcolor{LightGreen}None    &   \textbf{ReFixMatch}                                  & \textbf{75.2}     & \textbf{91.9}           & 25.6M/25.6M   &  $\sim400$     \\
    % MoCo-EMAN\cite{cai2021exponential}    &   ReFixMatch                                  & \textbf{75.2}     & \textbf{91.9}           & 25.6M/25.6M   &  $\sim400$     \\ 
    \midrule    \bottomrule
    \end{tabular}%
\end{sc}
\end{table*}

\section{List of Data Transformations}
We report the detailed augmentations used in our method in Table \ref{tab:randaugment}.
This list of transformations is similar to the original list used in FixMatch \cite{sohn2020fixmatch} and FlexMatch \cite{zhang2021flexmatch}.

\begin{table*}[!ht]
    \centering
    \caption{List of transformations used in RandAugment}
    \label{tab:randaugment}
\begin{sc}
    \begin{tabular}{p{2.5cm}p{10cm}p{1.5cm}p{1.5cm}}
    \toprule    \midrule
    Transformation & Description & Parameter & Range\\
    \midrule    \midrule
    Autocontrast & Maximizes the image contrast by setting the darkest (lightest) pixel to black (white). & &\\
    Brightness & Adjusts the brightness of the image. $B = 0$ returns a black
    image, $B = 1$ returns the original image. & $B$ & [0.05, 0.95]\\
    Color & Adjusts the color balance of the image like in a TV. $C = 0$ returns a black \& white image, $C = 1$ returns the original image. & $C$ & [0.05, 0.95]\\
    Contrast & Controls the contrast of the image. A $C = 0$ returns a gray image, $C = 1$ returns the original image. & $C$ & [0.05, 0.95]\\
    Equalize & Equalizes the image histogram. & &\\
    Identity & Returns the original image. & &\\
    Posterize & Reduces each pixel to $B$ bits. & $B$ & [4, 8]\\
    Rotate & Rotates the image by $\theta$ degrees. & $\theta$ & [-30, 30]\\
    Sharpness & Adjusts the sharpness of the image, where $S = 0$ returns a blurred image, and $S = 1$ returns the original image. & $S$ & [0.05, 0.95]\\
    Shear\_x & Shears the image along the horizontal axis with rate $R$. & $R$ & [-0.3, 0.3] \\
    Shear\_y & Shears the image along the vertical axis with rate $R$. & $R$ & [-0.3, 0.3]\\
    Solarize & Inverts all pixels above a threshold value of $T$. & $T$ & [0, 1]\\
    Translate\_x & Translates the image horizontally by ($\lambda\times$image width) pixels. & $\lambda$ & [-0.3, 0.3]\\
    Translate\_y & Translates the image vertically by ($\lambda\times$image height) pixels. & $\lambda$ & [-0.3, 0.3] \\
    \midrule    \bottomrule
    \end{tabular}
\end{sc}
\end{table*}

\section*{Precision, Recall, F1 and AUC}
We further report precision, recall, F1-score, and AUC (area under curve) results on the CIFAR-10 dataset.
As shown in Table \ref{table:details2}, ReFixMatch also has the best performance on precision, recall, F1-score, and AUC.

\begin{table*}[ht!]
\centering
    \caption{Precision, recall, F1-score and AUC results on CIFAR-10.}
    \label{table:details2}
% \resizebox{0.7\linewidth}{!}{
\begin{sc}
    \begin{tabular}{@{}l|cccc|cccc}
    \toprule    \midrule
    Label Amount    & \multicolumn{4}{c}{40 labels}      & \multicolumn{4}{|c}{4000 labels}  \\ \midrule
    Criteria    & Precision & Recall    & F1-score  & AUC       & Precision & Recall    & F1-score   & AUC \\ \midrule  \midrule
    FixMatch    & 0.9333    & 0.9290    & 0.9278    & 0.9910    & 0.9571    & 0.9571    & 0.9569  & 0.9984      \\ \midrule
    FlexMatch   & 0.9506    & 0.9507    & 0.9506    & 0.9975    & 0.9580    & 0.9581    & 0.9580  & 0.9984       \\ \midrule
    \rowcolor{LightGreen}\textbf{ReFixMatch}   & \textbf{0.9513}   & \textbf{0.9513}    & \textbf{0.9510}    & \textbf{0.9976}    & \textbf{0.9582}    & \textbf{0.9583}    & \textbf{0.9582}  & \textbf{0.9986}        \\ \midrule   \bottomrule
    \end{tabular}%
\end{sc}
% }
\end{table*}

% \section*{Limitation}
% ReFixMatch has shown the ability to improve the general performance of the existing SSL framework by leveraging the whole unlabeled dataset without introducing overhead.
% However, this ability somewhat conflicts with other methods using flexible thresholds, such as CPL.
% How to effectively and efficiently integrate ReFixMatch with the other methods using flexible thresholds is a potential research direction.

%%%%%%%%%%%%%%%%%%%%%%%%%%%%%%%%%%%%%%%%%%%%%%%%%%%%%%%%%%%%%%%%%%%%%%%%%%%%%%%
%%%%%%%%%%%%%%%%%%%%%%%%%%%%%%%%%%%%%%%%%%%%%%%%%%%%%%%%%%%%%%%%%%%%%%%%%%%%%%%
{\small
\bibliographystyle{ieee_fullname}
\bibliography{example_paper}
}

\end{document}